\documentclass{article}
\pdfoutput=1
\usepackage[numbers]{natbib}

\usepackage[final]{nips_2018}

\usepackage[utf8]{inputenc} % allow utf-8 input
\usepackage[T1]{fontenc}    % use 8-bit T1 fonts
\usepackage{hyperref}       % hyperlinks
\usepackage{url}            % simple URL typesetting
\usepackage{booktabs}       % professional-quality tables
\usepackage{amsfonts}       % blackboard math symbols
\usepackage{nicefrac}       % compact symbols for 1/2, etc.
\usepackage{microtype}      % microtypography
\usepackage{algorithm,algorithmic}
\usepackage{amsmath}
\usepackage[pdftex]{graphicx}

\usepackage{amsthm}
\usepackage{enumitem}
\usepackage{multirow}
\usepackage[dvipsnames]{xcolor}

\definecolor{mygray}{gray}{0.5}

\title{Modeling Attention Flow on Graphs}

\author{
  Xiaoran Xu$^1$,\; Songpeng Zu$^1$,\; Chengliang Gao$^2$\thanks{Work done during the internship in Hulu},\; Yuan Zhang$^2$\footnotemark[1],\; Wei Feng$^1$ \\
  $^1$Hulu Innovation Lab, Beijing, China \\
  \texttt{\{xiaoran.xu, songpeng.zu, wei.feng\}@hulu.com} \\
  $^2$School of Electronics Engineering and Computer Science, Peking University, Beijing, China \\
  \texttt{\{gaochengliang, yuan.z\}@pku.edu.cn} \\
}
 
\begin{document}

\maketitle

\begin{abstract}
Real-world scenarios demand reasoning about process, more than final outcome prediction, to discover latent causal chains and better understand complex systems. It requires the learning algorithms to offer both accurate predictions and clear interpretations. We design a set of trajectory reasoning tasks on graphs with only the source and the destination observed. We present the \textit{attention flow mechanism} to explicitly model the reasoning process, leveraging the relational inductive biases by basing our models on graph networks. We study the way attention flow can effectively act on the underlying information flow implemented by message passing. Experiments demonstrate that the attention flow driven by and interacting with graph networks can provide higher accuracy in prediction and better interpretation for trajectory reasoning.
\end{abstract}
\section{Introduction}
Many practical applications have the need to infer latent causal chains or to construct interpretations for observations or some predicted results. For example, in a physical world, we want to reason the trajectories of moving objects given very few observed frames; in a video streaming system, we wish recommendation models that track the evolving user interests to provide personalized recommendation reasons linking users' watched videos to the recommended ones. Here, we focus on graph-based scenarios, and aim to infer latent chains that might cause observed results described by nodes and edges in a given graph.

Graph networks \cite{gori2005new,scarselli2009graph,battaglia2018relational,sanchez2018graph,hamrick2018relational} are a family of neural networks that operate on graphs and carry strong relational inductive biases. It is believed that graph networks have a powerful fitting capacity to deal with graph-structured data. However, its black-box nature makes it less competitive than other differentiable logic-based reasoning \cite{cohen2016tensorlog,cohen2017tensorlog,yang2017differentiable} when modeling the reasoning process with interpretations provided. In this work, we develop a new attention mechanism on graphs, called \textit{attention flow}, to model the reasoning process to predict the final outcome with the interpretability. We use the message passing algorithm in graph networks to derive a transition matrix evolving with time steps to drive the attention flow. We also let the attention flow act back on the passed messages, called \textit{information flow}. To evaluate the models, we design a set of trajectory reasoning tasks, where only the source and destination ends of trajectories are observed.

Our contributions are two-fold. First, our attention flow mechanism, built on graph networks, introduces a new way to construct interpretations and increase the transparency when applying graph networks. Second, we show how attention flow can effectively intervene back in message passing conducted in graph networks, analogous to that of reinforcement learning where actions taken by agents would affect states of the environment. Experiments demonstrate that the graph network models with the explicit and backward-acting attention flow compare favorably both in prediction accuracy and in interpretability against those without it.

\textbf{Related Works}. Graph networks\cite{battaglia2018relational,sanchez2018graph,hamrick2018relational}, dating back to a decade ago \cite{gori2005new,scarselli2009graph}, are thought to support relational reasoning and combinatorial generalization over graph-structured representations. Recently, this area has grown rapidly and many versions of graph networks have been proposed, including Gated Graph Neural Networks \cite{li2015gated}, Interaction Networks \cite{battaglia2016interaction}, Relation Networks \cite{raposo2017discovering}, Message Passing Neural Networks \cite{gilmer2017neural}, Graph Attention Networks \cite{velickovic2017graph}, Non-Local Neural Networks \cite{wang2018non}, and the graph convolutional network family \cite{bronstein2017geometric}, spectral \cite{bruna2013spectral,henaff2015deep,defferrard2016convolutional,kipf2016semi} or non-spectral \cite{duvenaud2015convolutional,niepert2016learning,monti2017geometric,atwood2016diffusion,hamilton2017inductive}. From a unified perspective, \cite{gilmer2017neural} introduces the message passing mechanism to generalize computation frameworks on graphs; \cite{battaglia2018relational} uses the term of \textit{graph networks} to generalize and extend several lines in this area. While graph networks give reasoning over graphs more fitting capacity, we look back on the old-fashioned logic and rules-based reasoning to seek the interpretability. Recent probabilistic logic programming, such as TensorLog \cite{cohen2016tensorlog,cohen2017tensorlog} and NeuralLP \cite{yang2017differentiable}, develops differentiable reasoning based on a knowledge graph, learning soft logic rules in an end-to-end style, and the process is much like a rooted random walk computing conditional probabilities based on paths. People also studied reasoning over paths or graphs using reinforcement learning to deal with discrete actions of choosing nodes or edges, such as MINERVA \cite{das2017go}, Structure2Vec Deep Q-learning \cite{khalil2017learning}, and Neural Combinatorial Optimization \cite{bello2016neural}. Attention mechanisms, derived from sequence-based tasks \cite{bahdanau2014neural}, developed in \cite{vaswani2017attention} referred to as \textit{self-attention}, have been brought in graphs recently by attending over neighbors of each node \cite{velickovic2017graph,kool2018attention} or non-local areas \cite{wang2018non}. Here, we present the attention flow mechanism not only for the computation need but also for the interpretation purpose. 
\section{Tasks}
\label{sec:tasks}
Real-world scenarios often demand reasoning about process, that is, constructing interpretations by listing a series of causal connections linking an opening to an outcome. We need a simulation system to generate a trajectory of events with the dynamics governed by latent factors, such as the trajectory of a moving object controlled by an external force. Instead of full observation, we only allow events at the source and the destination observed, treating the task as a trajectory reasoning problem.

We build a corrupted $N \times N$ grid world with a small fraction of nodes or edges randomly removed. There are 8 types of directed edges at most for each node, connecting it to its neighbors, such as \textit{east (E)} and \textit{northeast (NE)}. Picking an arbitrary node as the source $v^0$, we draw a sequence of consecutive nodes to construct a trajectory $(v^0, v^1, \ldots, v^T)$ and obtain the final node $v^T$ as the destination. Each node on the trajectory except the source is chosen from the neighborhood of the previous node $v_{x,y}$ by drawing one of the 8 edge types $e$ from the distribution below driven by latent direction function $\vec{d}_{t,x,y}$ varying with time $t$ and location $(x,y)$: 
\begin{equation}
\begin{split}
& P(e) \propto \exp{\Big( \big\langle \vec{d}_{e}/\|\vec{d}_{e}\|, \vec{d}_{t,x,y}/\|\vec{d}_{t,x,y}\| \big\rangle / \sigma^2 \Big)},\quad e\in \{E,NE,N,NW,W,SW,S,SE\} \\
& \vec{d}_{t,x,y} =(a_1 \cos{\theta_{t,x,y}} + b_1, a_2 \sin{\theta_{t,x,y}} + b_2),\quad
\theta_{t,x,y} = \omega t + \lambda_1 x + \lambda_2 y + \phi
\end{split}
\end{equation}
The trajectory terminates by either choosing a non-existent edge or reaching the maximal steps. To be specific, we generate four types of trajectories governed by: 
\begin{itemize}[nosep, wide=0pt, leftmargin=\dimexpr\labelwidth + 2\labelsep\relax]
    \item $\vec{d}_{t,x,y}=(1,0.4)$, a straight line with a slope.
    \item $\vec{d}_{t,x,y}=(1, \sin(0.4 t + 1.6))$, a sine curve with directions varying with time $t$.
    \item $\vec{d}_{t,x,y}=(\cos \theta_{x,y}, \sin \theta_{x,y})$, $\theta_{x,y}=0.2 x + 0.2 y$, directions varying with the current location.
    \item $\vec{d}_{t,x,y}=(\cos \theta_{x,y}, \sin \theta_{x,y})$, $\theta_{x,y}=0.2 \max{\{x_i\}} + 0.2 \max{\{y_i\}}$, depending on location history.
\end{itemize}

Instead of learning a latent model to solve the trajectory reasoning problem, we use a supervised setting. Considering only the source and the destination available, we train a discriminative model to predict the destination node by inputting the source node. We leverage the graph structure in the corrupted grid world, as the problem implies strong inductive biases on graphs, relational (\textit{sequences of consecutive nodes}) and non-relational (\textit{latent direction functions depending on time, location and history}). The trajectory reasoning problem is difficult considering that many candidate paths link the source to the destination. The only clue we can observe is the destination nodes resulting from the blocked trajectories caused by removed nodes or edges. Note that we do not look for the shortest paths but the true trajectory pattern governed by some latent dynamics. The evaluation criteria should be based on both the accuracy of prediction and the human readability of interpretation.

\section{Models}
\textbf{Modeling attention flow on graphs}. We view the problem of predicting the destination given the source as predicting the output attention distribution over nodes given the input attention distribution. We use the term of attention distribution to represent the probability distribution of attending over nodes. For each pair $(v_{src}, v_{dst})$, the input attention distribution has all the probability mass concentrated on the source node. After a series of computation on the graph, the resulting output attention distribution predicts the most likely node to be the destination. Attention transfers from the source to the destination, implying a flow through the graph mimicking latent causal chains. The followings attempt to model the attention flow on graphs from three different perspectives.

\textbf{Implicit attention flow in graph networks}. Modern graph networks (GNs) mostly employ the message passing mechanism implemented by neural network building blocks, such as MLP and GRU modules. Representations in GNs include node-level states $\mathbf{h}_i\in \mathbb{R}^d$, edge-level messages $\mathbf{m}_{ij}\in \mathbb{R}^d$ and sometimes graph-level global state $\mathbf{g}\in \mathbb{R}^d$. A GN framework has three phases: the initialization phase, the propagation phase, and the output phase. The model in the propagation phase includes:
\begin{itemize}[nosep, wide=0pt, leftmargin=\dimexpr\labelwidth + 2\labelsep\relax]
    \item Message function: $\mathbf{m}^t_{ij} = f_{\mathrm{msg}}(\mathbf{h}^t_i, \mathbf{h}^t_j, \mathbf{g}^t; \theta_{e_{ij}})$, where $\theta_{e_{ij}}$ is edge type-specific parameters.
    \item Message aggregation operation: $\mathbf{\bar{m}}^t_j = \sum_i \mathbf{m}^t_{ij}$, aggregating all received messages from neighbors.
    \item Node update function: $\mathbf{h}^{t+1}_i = f_{\mathrm{node}}(\mathbf{h}^t_i, \mathbf{\bar{m}}^t_i, \mathbf{g}^t, \mathbf{u}_i)$, where $\mathbf{u}_i$ is stationary node embeddings.
    \item Global update function: $\mathbf{g}^{t+1} = f_{\mathrm{global}}(\mathbf{g}^t, \mathbf{\bar{h}}^t, \mathbf{\bar{m}}^t)$, where $\mathbf{\bar{h}}^t = \frac{1}{|\mathcal{V}|}\sum_i \mathbf{h}^t_i, \; \mathbf{\bar{m}}^t = \frac{1}{|\mathcal{V}|}\sum_i \mathbf{\bar{m}}^t_i$.
\end{itemize}
To model the attention flow, we modify the initialization phase by defining $\mathbf{h}^0_i:=[\mathbf{\dot{h}}^0_i:\mathbf{\ddot{h}}^0_i]$ where $\mathbf{\dot{h}}^0_i\in \mathbb{R}^{d'}$ is attention channels and $\mathbf{\ddot{h}}^0_i\in \mathbb{R}^{d-d'}$ auxiliary channels. We initialize $\mathbf{\dot{h}}^0_\mathrm{src}=\mathbf{\tilde{1}}$ for the source and $\mathbf{\dot{h}}^0_{i}=\mathbf{0}$ for the rest where $\mathbf{\tilde{1}}:=\mathbf{1}/\|\mathbf{1}\|$ acts as a reference vector for computing attention distributions, so that score $\langle \mathbf{\dot{h}}^0_i, \mathbf{\tilde{1}} \rangle$ is $1$ on the source node and $0$ on the rest. We set $\mathbf{\ddot{h}}^0_i = f_{\mathrm{init}}(\mathbf{u}_i)$. At the output phase, we compute the output attention distribution by $\mathrm{softmax}({\langle \mathbf{\dot{h}}^T_i, \mathbf{\tilde{1}} \rangle}_{i=1}^n)$.

This model wraps the attention flow in the messages passing process at the beginning and takes it out at the end. Neural network-based computation makes the propagation model a black box, lacking an explicit way to depict the attention flow, helpless for the interpretation purpose.

\textbf{Explicit attention flow by random walks}. To explicitly model the attention flow, we use random walks with learnable transition $\mathbf{T}$. The dynamics of attention flow are driven by $\mathbf{a}^{t+1}=\mathbf{T}\mathbf{a}^t$ where $\mathbf{a}^{t+1}$ and $\mathbf{a}^t$ represent two consecutive attention distributions. Here, we take two model settings: 
\begin{itemize}[nosep, wide=0pt, leftmargin=\dimexpr\labelwidth + 2\labelsep\relax]
    \item Stationary transition setting: $\tau_{ij} = f_\mathrm{trans}(\mathbf{u}_i, \mathbf{u}_j; \phi_{e_{ij}})$ and do the row-level softmax to get entry $T_{ij}$. The transition $\mathbf{T}$ is stationary across inputs and steps. 
    \item Dynamic transition setting: $\tau^t_{ij} = f_\mathrm{trans}(\mathbf{h}^t_i, \mathbf{h}^t_j; \phi_{e_{ij}})$ where $\mathbf{h}^{t+1}_i = f_{\mathrm{node}}(\mathbf{h}^t_i, \mathbf{g}^t, \mathbf{u}_i)$, $\mathbf{g}^{t+1} = f_{\mathrm{global}}(\mathbf{g}^t, \mathbf{\bar{h}}^t)$ and $\mathbf{\bar{h}}^t = \sum_i a^t_i \mathbf{h}^t_i$. Note that no message passing is applied. The update on global state $\mathbf{g}^t$ is based on the weighted sum of node states affected by $\mathbf{a}^t$. We emphasize that attention distributions can act as more than output of internal states and effectively impact back on the internal. The graph context captured this way is still limited without leveraging message passing.
\end{itemize}

\begin{figure*}[t]
\centering
\includegraphics[width=0.95\textwidth]{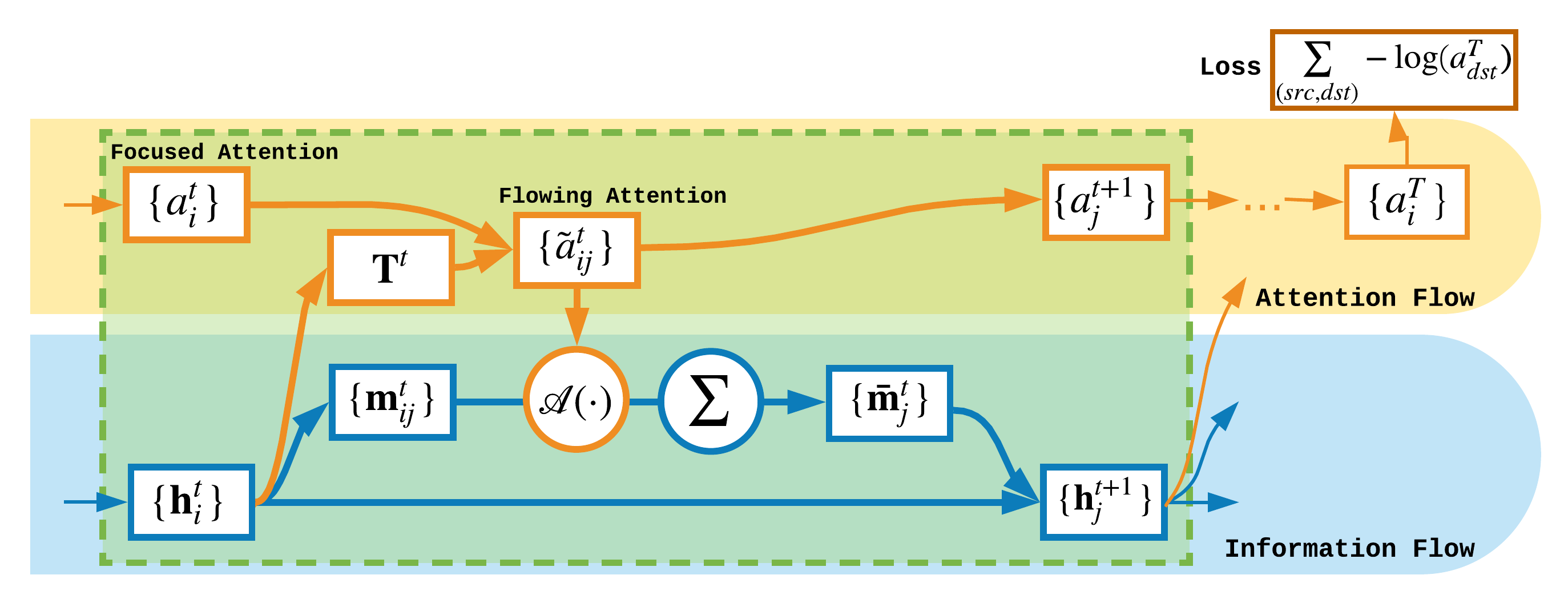}
\vspace{-5pt}
\caption{\footnotesize{The two-flow model architecture.}}
\vspace{-10pt}
\label{fig:model}
\end{figure*}

\textbf{Explicit attention flow with graph networks}. The interpretability benefit of explicit attention flow can be gained even when enjoying the expressivity of graph networks. We present the attention flow mechanism by introducing the node-level attention, called \textit{focused attention} $\mathbf{a}$, and the edge-level attention, called \textit{flowing attention} $\mathbf{\tilde{a}}$. With the superscript step $t$, the dynamics are driven by:
\begin{equation}
\tilde{a}^t_{ij} = T^t_{ij} a^t_i, \quad a^{t+1}_j = \sum_i \tilde{a}^t_{ij}, \quad \mathrm{ s.t. } \sum_i a^t_i=1, \quad \sum_{ij} \tilde{a}^t_{ij}=1,
\end{equation}
where transition $\mathbf{T}^t$ relies on the rich context carried by the underlying message passing in GNs, thus called \textit{information flow}, in addition to attention flow. See the two-flow model architecture in Figure \ref{fig:model}.

It is obvious that the information flow determines the attention flow, but we are more curious about how the attention flow can affect back the information flow. We have seen the node-level backward acting of the focused attention on the sum of node states as above. Here, we study on the edge level how the flowing attention acts on the information flow by defining the message-attending function $\mathbf{\tilde{m}}^t_{ij} = \mathcal{A}(\tilde{a}^t_{ij}, \mathbf{m}^t_{ij})$ to produce attended message $\mathbf{\tilde{m}}^t_{ij}$ to replace original message $\mathbf{m}^t_{ij}$:
\begin{itemize}[nosep, wide=0pt, leftmargin=\dimexpr\labelwidth + 2\labelsep\relax]
    \item No acting: $\mathcal{A}(\tilde{a}^t_{ij}, \mathbf{m}^t_{ij}):=\mathbf{m}^t_{ij}$
    \item Multiplying: $\mathcal{A}(\tilde{a}^t_{ij}, \mathbf{m}^t_{ij}):=\tilde{a}^t_{ij} \cdot \mathbf{m}^t_{ij}$
    \item Non-linear acting after multiplying: $\mathcal{A}(\tilde{a}^t_{ij}, \mathbf{m}^t_{ij}):=\mathrm{MLP}(\tilde{a}^t_{ij} \cdot \mathbf{m}^t_{ij})$
\end{itemize}
To design a meaningful $\mathcal{A}(\tilde{a}^t_{ij}, \mathbf{m}^t_{ij})$, when $\tilde{a}^t_{ij}=0$ we make $\mathcal{A}(0, \mathbf{m}^t_{ij})$ independent from $\mathbf{m}^t_{ij}$, implying no attention paid to this piece of message but not necessarily being $0$. We find $\mathrm{MLP}(\tilde{a}^t_{ij} \cdot \mathbf{m}^t_{ij})$ performs the best in most cases, revealing not only the importance of backward acting but also the necessity of keeping information flow even if not being attended.

\textbf{Connections to reinforcement learning and probabilistic latent models}. If we inject noises and then pick the top-1 attended node each step, the process becomes similar to reinforcement learning in some way. If we apply noises but keep it soft in the Gumbel-Softmax \cite{Jang2017_GumbelSoftmax} or Concrete \cite{Maddison2017_Concrete} distribution, it turns into a probabilistic latent model. Attention flow can be viewed as graph-level computation operating directly andnumerically in a probability space rather than in a discrete sample space.

\section{Experiments}

\subsection{Experimental Procedure}

\textbf{Dataset generation and statistics}. We generate a number of dataset groups each representing a randomized grid world with a specific version of trajectories, consisting of sequences of nodes, driven by a specific setting of latent dynamics applied. More specifically, for each dataset group, we build a corrupted $N \times N$ grid world and then apply a latent direction function $\vec{d}_{t,x,y}$ to draw multiple trajectories starting from each node. The data generation parameters include:
\begin{itemize}[nosep, wide=0pt, leftmargin=\dimexpr\labelwidth + 2\labelsep\relax]
    \item $N$: The size of a grid map. Without nodes and edges dropped, we have $N^2$ nodes and $2(2N-1)(2N-2)$ directed edges at most, where each internal node is connected to 8 incoming edges as well as 8 outgoing edges. In the experiments, we test models in two sizes: $32 \times 32$ and $64 \times 64$.
    \item $p_\mathrm{node\_drp}$ and $p_\mathrm{edge\_drp}$: The dropping probabilities of randomly removing a node or an edge. If a node is removed, all its connected edges should be gone; if a node is left with no edges, that is, a single isolated point, we remove it. When we remove an edge connected by a pair of nodes $v_1$ and $v_2$, we drop edges $(v_1, v_2)$ and $(v_2, v_1)$ in both directions. In the experiments, we try two settings: dropping nodes only with $p_\mathrm{node\_drp}=0.1$, and dropping edges only with $p_\mathrm{edge\_drp}=0.2$.
    \item $T$: The maximal steps of a trajectory. Without being blocked, a trajectory would end with a maximal length of $T$. In the experiments, we set $T=16$ in $32 \times 32$ and $T=32$ in $64 \times 64$. Therefore, the total problem scale depends on $N$, $p_\mathrm{node\_drp}$ and $p_\mathrm{edge\_drp}$, and $T$.
    \item $\sigma$: The standard deviation of sampling an edge around latent direction $\vec{d}_{t,x,y}$. Larger $\sigma$ increases the chance to bypass gaps caused by removed nodes and edges, and also brings a larger deviation of positions of the destination nodes given the same source, leading to a larger exploring area and more uncertainty for prediction. In the experiments, we pick two values, $\sigma=0.2$ and $\sigma=0.5$. 
    \item $n_\mathrm{rollout}$: The number of rollouts to draw a trajectory starting from a specific node. With each node as a source, we try $n_\mathrm{rollout}$ times and then remove the duplicated source-destination pairs. In the experiments, we use $n_\mathrm{rollout}=10$.
    \item $\vec{d}_{t,x,y}$: The latent direction function. In the experiments, we try four settings as shown in Section \ref{sec:tasks}: (1) a straight line with a constant sloped direction; (2) a sine curve with time-dependent varying directions; (3) a curve with location-dependent varying directions; (4) a curve with location history-dependent varying directions.
    \item $seed$: The datasets in each group share the same generation parameters listed above except $seed$. The purpose is to make experimental results less impacted by accidental factors. We use five random seeds in the experiments.
\end{itemize}

We generate $24$ dataset groups with their names and generation parameters listed in Table \ref{tab:params-gen} in the appendix. Each dataset group contains five datasets with different $seed$s. The observed part of each dataset includes a grid map containing all edge information and a list of source-destination pairs used for training, validation and test. We make the training, validation and test sets by an $8:1:1$ splitting on source nodes, so that we can assess models based on their performances of handling pairs with unseen source nodes. The statistics of datasets are given in Table \ref{tab:dataset-stats} in the appendix.

\textbf{Models in comparison}. To fully evaluate the attention flow mechanism, we choose three types of graph networks plus two random walk-based models to set the benchmark. Graph networks include a full Graph Network (FullGN) \cite{battaglia2018relational}, a Gated Graph Neural Network (GGNN) \cite{li2015gated}, and a Graph Attention Network (GAT) \cite{velickovic2017graph}. Note that the regular versions of these graph networks are incapable to explicitly model attention flow and fulfill the trajectory reasoning purpose, though able to make predictions. We remodel them with the attention flow mechanism imposed respectively, which is implemented in three different ways on how the flowing attention acts back on the passed messages.

\textit{Regular graph networks}. Node states $\mathbf{h}^t_i=[\mathbf{\dot{h}}^t_i : \mathbf{\ddot{h}}^t_i] \in \mathbb{R}^d$ have $d=40$ channels (or dimensions) where the number of attention channels is $d' = 8$ and the rest channels are left for carrying auxiliary messages. We tried several combinations for the pair of channel numbers and found 8 attention channels performed the best. To initialize $\mathbf{\dot{h}}^0_i$, we set $\mathbf{\dot{h}}^0_{src} = \mathbf{\tilde{1}} = \mathbf{1}/\sqrt{d'}$ for the source node and $\mathbf{\dot{h}}^0_i = \mathbf{0}$ for the rest. We also initialize $\mathbf{\ddot{h}}^0_i = f_{\mathrm{init}}(\mathbf{u}_i)$ where $\mathbf{u}_i\in \mathbb{R}^d$ represents stationary node embeddings and $f_{\mathrm{init}}$ is a single-layer feedforward network with the $\mathrm{tanh}$ activation function. The loss is defined in the cross entropy between the one-hot true destination labels and the predicted probability distribution by $\mathrm{softmax}({\langle \mathbf{\dot{h}}^T_i, \mathbf{\tilde{1}} \rangle}_{i=1}^n)$.
\begin{itemize}[nosep, wide=0pt, leftmargin=\dimexpr\labelwidth + 2\labelsep\relax]
    \item \textit{FullGN}: This model has global state $\mathbf{g}^t$, and both the message function and the node update function take $\mathbf{g}^t$ as one of their inputs. Here, each function uses a single-layer feedforward network.
        \begin{itemize}[nosep, wide=0pt, leftmargin=\dimexpr\labelwidth + 2\labelsep\relax]
            \item Message function: $\mathbf{m}^t_{ij} = \tanh(\mathbf{W}_{e_{ij}}[\mathbf{h}^t_i : \mathbf{h}^t_j : \mathbf{g}^t] + \mathbf{b}_{e_{ij}})$.
            \item Node update function: $\mathbf{h}^{t+1}_i = \tanh(\mathbf{W}_\mathrm{node}[\mathbf{h}^t_i : \mathbf{\bar{m}}^t_i : \mathbf{u}_i : \mathbf{g}^t] + \mathbf{b}_\mathrm{node})$.
            \item Global update function: $\mathbf{g}^{t+1} = \tanh(\mathbf{W}_\mathrm{global}[\mathbf{g}^t : \mathbf{\bar{h}}^t : \mathbf{\bar{m}}^t] + \mathbf{b}_\mathrm{global})$.
        \end{itemize}
    \item \textit{GGNN}: This model computes messages in a non-pairwise linear manner that depends on sending nodes and edge types. It ignores the global state and changes the node update function into a gated recurrent unit (GRU).
        \begin{itemize}[nosep, wide=0pt, leftmargin=\dimexpr\labelwidth + 2\labelsep\relax]
            \item Message function: $\mathbf{m}^t_{ij} = \mathbf{W}_{e_{ij}} \mathbf{h}^t_i  + \mathbf{b}_{e_{ij}}$.
            \item Node update function: $\mathbf{h}^{t+1}_i = \mathrm{GRU}(\mathbf{h}^t_i,\; [\mathbf{\bar{m}}^t_i : \mathbf{u}_i])$.
        \end{itemize}
    \item \textit{GAT}: This model uses multi-head self-attention layers. Here, we take edge types into account to define weights. We also apply a GRU to the node update function, taking the concatenation of all multi-head aggregated messages $\mathbf{\bar{m}}_i^{t,k}$ and node embedding $\mathbf{u}_i$ as the input. We use $K=5$ heads each with an $8$-dimensional self-attention so that the concatenated message still has $40$ dimensions.
        \begin{itemize}[nosep, wide=0pt, leftmargin=\dimexpr\labelwidth + 2\labelsep\relax]
            \item Multi-head self-attention: $\alpha^{t,k}_{ij} = \mathrm{softmax}_j(\mathrm{LeakyRelu}( \mathbf{a}^T [\mathbf{W}^k_{e_{i*}} \mathbf{h}^t_i : \mathbf{W}^k_{e_{i*}} \mathbf{h}^t_*]))$.
            \item Message function: $\mathbf{m}^{t,k}_{ij} = \alpha^{t,k}_{ij} \mathbf{W}^k_{e_{ij}} \mathbf{h}^t_i$.
            \item Node update function: $\mathbf{h}^{t+1}_i = \mathrm{GRU}(\mathbf{h}^t_i,\; [\mathbf{\bar{m}}^{t,1}_i : \ldots : \mathbf{\bar{m}}^{t,K}_i : \mathbf{u}_i])$.
        \end{itemize}
\end{itemize}

\textit{Remodeled graph networks with explicit attention flow}. We add our attention flow module onto the computation framework of graph networks as shown in Figure \ref{fig:model}. At the initialization phase, we define $\mathbf{a}^0$ by $a_{src}^0=1$ and $a_i^0=0$ for the rest; at the output phase, we take $\mathbf{a}^T$ to compute the loss:
$$\sum_{(src,dst)} - \log a_{dst}^T$$
For the propagation phase, we need to compute two more functions:
\begin{itemize}[nosep, wide=0pt, leftmargin=\dimexpr\labelwidth + 2\labelsep\relax]
    \item Transition logits function: $\tau^t_{ij} = \mathbf{w}^T_{e_{ij}} [\mathbf{h}^t_i : \mathbf{h}^t_j : (\mathbf{h}^t_i \otimes \mathbf{h}^t_j)] + b_{e_{ij}}$ in order to compute $\mathbf{T}^t$.
    \item Message-attending function: $\mathbf{\tilde{m}}^t_{ij} = \mathcal{A}(\tilde{a}^t_{ij}, \mathbf{m}^t_{ij})$ to produce attended message $\mathbf{\tilde{m}}^t_{ij}$ in place of original message $\mathbf{m}^t_{ij}$. We study three ways to implement it: (1) no acting, (2) multiplying, (3) non-linear acting after multiplying. For (3), we simply use a single-layer feedforward network. Finally, we derive three explicit attention flow models based on each graph network.
    \begin{itemize}[nosep, wide=0pt, leftmargin=\dimexpr\labelwidth + 2\labelsep\relax]
        \item \textit{\{FullGN, GGNN, GAT\}-NoAct}.
        \item \textit{\{FullGN, GGNN, GAT\}-Mul}.
        \item \textit{\{FullGN, GGNN, GAT\}-MulMlp}.
    \end{itemize}
\end{itemize}

\textit{Random walk-based models}. If we model the attention flow without considering message passing conducted in graph networks, the method falls into the family of differentiable random walk models with a learned transition matrix. Here, we try two types of transition, stationary and dynamic.
\begin{itemize}[nosep, wide=0pt, leftmargin=\dimexpr\labelwidth + 2\labelsep\relax]
    \item \textit{RW-Stationary}:
        \begin{itemize}[nosep, wide=0pt, leftmargin=\dimexpr\labelwidth + 2\labelsep\relax]
            \item Transition logits function: $\tau^t_{ij} = \mathbf{w}^T_{e_{ij}} [\mathbf{u}_i : \mathbf{u}_j : (\mathbf{u}_i \otimes \mathbf{u}_j)] + b_{e_{ij}}$
        \end{itemize}
    \item \textit{RW-Dynamic}:
        \begin{itemize}[nosep, wide=0pt, leftmargin=\dimexpr\labelwidth + 2\labelsep\relax]
            \item Transition logits function: $\tau^t_{ij} = \mathbf{w}^T_{e_{ij}} [\mathbf{h}^t_i : \mathbf{h}^t_j : (\mathbf{h}^t_i \otimes \mathbf{h}^t_j)] + b_{e_{ij}}$.
            \item Node update function: $\mathbf{h}^{t+1}_i = \tanh(\mathbf{W}_\mathrm{node}[\mathbf{h}^t_i : \mathbf{u}_i : \mathbf{g}^t] + \mathbf{b}_\mathrm{node})$.
            \item Global update function: $\mathbf{g}^{t+1} = \tanh(\mathbf{W}_\mathrm{global}[\mathbf{g}^t : \mathbf{\bar{h}}^t] + \mathbf{b}_\mathrm{global})$ where $\mathbf{\bar{h}}^t = \sum_i a^t_i \mathbf{h}^t_i$.
        \end{itemize}
\end{itemize}

\begin{table}[t]
  \caption{\footnotesize{Comparison results on dataset groups \textit{\{LINE,SINE,LOCATION,HISTORY\}-SZ32-STP16-NDRP-STD0.2}. This table focuses on comparative evaluation between all the explicit attention flow models that offer clear interpretations as well as prediction results, so that we gray the results from the implicit attention flow models, that is, regular graph networks. Each column indicates one comparison in a specific metric based on the same dataset group. $*$ represents the highest metric score acquired by random walk-based models. $\checkmark$ represents the graph network-based explicit attention flow models that beat the best random walk-based models. Futhermore, we compare the three message-attending approaches between the explicit attention flow models based on the same graph network, and then highlight the best in bold.}}
  \label{tab:comparison-results-std0.2}
  \centering
  \begin{tabular}{l|cc|cc|cc|cc}
    \noalign{\hrule height 1.2pt}
    & \multicolumn{2}{c|}{\footnotesize{LINE}} & \multicolumn{2}{|c|}{\footnotesize{SINE}} & \multicolumn{2}{|c|}{\footnotesize{LOCATION}} & \multicolumn{2}{|c}{\footnotesize{HISTORY}} \\
    Model & \footnotesize{H@1(\%)} & \footnotesize{MRR} & \footnotesize{H@1(\%)} & \footnotesize{MRR} & \footnotesize{H@1(\%)} & \footnotesize{MRR} & \footnotesize{H@1(\%)} & \footnotesize{MRR} \\
    \noalign{\hrule height 1.2pt}
    \footnotesize{RW-Stationary} & \footnotesize{15.80}$\enspace$ & \footnotesize{0.3409}$\enspace$ & \footnotesize{8.56} & \footnotesize{0.2177}$\enspace$ & \footnotesize{50.06}\scriptsize{$*$} & \footnotesize{0.6625}\scriptsize{$*$} & \footnotesize{16.44}$\enspace$ & \footnotesize{0.3215}$\enspace$ \\
    \footnotesize{RW-Dynamic} & \footnotesize{16.64}\scriptsize{$*$} & \footnotesize{0.3562}\scriptsize{$*$} & \footnotesize{19.15}\scriptsize{$*$} & \footnotesize{0.3418}\scriptsize{$*$} & \footnotesize{45.94}$\enspace$ & \footnotesize{0.6320}$\enspace$ & \footnotesize{20.41}\scriptsize{$*$} & \footnotesize{0.3656}\scriptsize{$*$} \\
    \noalign{\hrule height 1.2pt}
    \footnotesize{FullGN} & \textcolor{mygray}{\textit{\footnotesize{15.13}}} & \textcolor{mygray}{\textit{\footnotesize{0.3451}}} & \textcolor{mygray}{\textit{\footnotesize{51.71}}} & \textcolor{mygray}{\textit{\footnotesize{0.6665}}} & \textcolor{mygray}{\textit{\footnotesize{25.67}}} & \textcolor{mygray}{\textit{\footnotesize{0.4393}}} & \textcolor{mygray}{\textit{\footnotesize{16.61}}} & \textcolor{mygray}{\textit{\footnotesize{0.3095}}} \\
    \cline{2-9}
    \footnotesize{FullGN-NoAct} & \footnotesize{16.65}\tiny{$\checkmark$} & \footnotesize{0.3574}\tiny{$\checkmark$} & \footnotesize{30.10}\tiny{$\checkmark$} & \footnotesize{0.4476}\tiny{$\checkmark$} & \footnotesize{46.47}$\;\:\,$ & \footnotesize{0.6337}$\;\:\,$ & \footnotesize{20.80}\tiny{$\checkmark$} & \footnotesize{0.3729}\tiny{$\checkmark$} \\
    \footnotesize{FullGN-Mul} & \footnotesize{16.69}\tiny{$\checkmark$} & \footnotesize{0.3636}\tiny{$\checkmark$} & \footnotesize{37.49}\tiny{$\checkmark$} & \footnotesize{0.4915}\tiny{$\checkmark$} & \footnotesize{43.44}$\;\:\,$ & \footnotesize{0.6029}$\;\:\,$ & \footnotesize{21.35}\tiny{$\checkmark$} & \footnotesize{0.3618}$\;\:\,$ \\
    \footnotesize{FullGN-MulMlp} & \textbf{\footnotesize{16.99}}\tiny{$\checkmark$} & \textbf{\footnotesize{0.3662}}\tiny{$\checkmark$} & \textbf{\footnotesize{39.91}}\tiny{$\checkmark$} & \textbf{\footnotesize{0.5195}}\tiny{$\checkmark$} & \textbf{\footnotesize{50.93}}\tiny{$\checkmark$} & \textbf{\footnotesize{0.6598}}$\;\:\,$ & \textbf{\footnotesize{23.94}}\tiny{$\checkmark$} & \textbf{\footnotesize{0.3850}}\tiny{$\checkmark$} \\
    \noalign{\hrule height 1.2pt}
    \footnotesize{GGNN} & \textcolor{mygray}{\textit{\footnotesize{15.49}}} & \textcolor{mygray}{\textit{\footnotesize{0.3493}}} & \textcolor{mygray}{\textit{\footnotesize{51.02}}} & \textcolor{mygray}{\textit{\footnotesize{0.6611}}} & \textcolor{mygray}{\textit{\footnotesize{29.20}}} & \textcolor{mygray}{\textit{\footnotesize{0.4699}}} & \textcolor{mygray}{\textit{\footnotesize{22.56}}} & \textcolor{mygray}{\textit{\footnotesize{0.3677}}} \\
    \cline{2-9}
    \footnotesize{GGNN-NoAct} & \footnotesize{16.64}$\;\:\,$ & \footnotesize{0.3570}\tiny{$\checkmark$} & \footnotesize{25.14}\tiny{$\checkmark$} & \footnotesize{0.3918}\tiny{$\checkmark$} & \footnotesize{49.50}$\;\:\,$ & \footnotesize{0.6621}$\;\:\,$ & \footnotesize{21.69}\tiny{$\checkmark$} & \footnotesize{0.3818}\tiny{$\checkmark$} \\
    \footnotesize{GGNN-Mul} & \footnotesize{16.95}\tiny{$\checkmark$} & \footnotesize{0.3610}\tiny{$\checkmark$} & \footnotesize{23.62}\tiny{$\checkmark$} & \footnotesize{0.3776}\tiny{$\checkmark$} & \footnotesize{45.22}$\;\:\,$ & \footnotesize{0.6185}$\;\:\,$ & \footnotesize{23.81}\tiny{$\checkmark$} & \footnotesize{0.3824}\tiny{$\checkmark$} \\
    \footnotesize{GGNN-MulMlp} & \textbf{\footnotesize{17.08}}\tiny{$\checkmark$} & \textbf{\footnotesize{0.3673}}\tiny{$\checkmark$} & \textbf{\footnotesize{34.75}}\tiny{$\checkmark$} & \textbf{\footnotesize{0.4699}}\tiny{$\checkmark$} & \textbf{\footnotesize{50.28}}\tiny{$\checkmark$} & \textbf{\footnotesize{0.6637}}\tiny{$\checkmark$} & \textbf{\footnotesize{26.06}}\tiny{$\checkmark$} & \textbf{\footnotesize{0.4001}}\tiny{$\checkmark$} \\
    \noalign{\hrule height 1.2pt}
    \footnotesize{GAT} & \textcolor{mygray}{\textit{\footnotesize{16.01}}} & \textcolor{mygray}{\textit{\footnotesize{0.3469}}} & \textcolor{mygray}{\textit{\footnotesize{43.19}}} & \textcolor{mygray}{\textit{\footnotesize{0.5566}}} & \textcolor{mygray}{\textit{\footnotesize{18.18}}} & \textcolor{mygray}{\textit{\footnotesize{0.3583}}} & \textcolor{mygray}{\textit{\footnotesize{12.11}}} & \textcolor{mygray}{\textit{\footnotesize{0.2333}}} \\
    \cline{2-9}
    \footnotesize{GAT-NoAct} & \footnotesize{16.02}$\;\:\,$ & \footnotesize{0.3536}$\;\:\,$ & \footnotesize{15.77}$\;\:\,$ & \footnotesize{0.3221}$\;\:\,$ & \footnotesize{46.10}$\;\:\,$ & \footnotesize{0.6356}$\;\:\,$ & \textbf{\footnotesize{23.17}}\tiny{$\checkmark$} & \textbf{\footnotesize{0.3818}}\tiny{$\checkmark$} \\
    \footnotesize{GAT-Mul} & \footnotesize{15.86}$\;\:\,$ & \footnotesize{0.3501}$\;\:\,$ & \footnotesize{20.14}\tiny{$\checkmark$} & \footnotesize{0.3429}\tiny{$\checkmark$} & \footnotesize{45.83}$\;\:\,$ & \footnotesize{0.6208}$\;\:\,$ & \footnotesize{22.70}\tiny{$\checkmark$} & \footnotesize{0.3762}\tiny{$\checkmark$} \\
    \footnotesize{GAT-MulMlp} & \textbf{\footnotesize{17.07}}\tiny{$\checkmark$} & \textbf{\footnotesize{0.3646}}\tiny{$\checkmark$} & \textbf{\footnotesize{30.64}}\tiny{$\checkmark$} & \textbf{\footnotesize{0.4390}}\tiny{$\checkmark$} & \textbf{\footnotesize{47.52}}$\;\:\,$ & \textbf{\footnotesize{0.6371}}$\;\:\,$ & \footnotesize{20.71}\tiny{$\checkmark$} & \footnotesize{0.3655}$\;\:\,$ \\
    \noalign{\hrule height 1.2pt}
  \end{tabular}
\end{table}

\textbf{Training and evaluation details}. Considering that trajectories might terminate before reaching the maximal steps, we add a selfloop edge onto each node so that we can treat all trajectories as ones with a fixed number of steps. Thus, there would be 9 types of edges during training. Our training hyperparameters include the batch size of $16$, the representation dimensions of $40$, the weight decay on node embeddings of $0.00001$, the decayed learning rates from $0.0005$ to $0.0001$ diminished by $0.0001$ every 10 epochs, and a total number of training epochs of $50$. We use the Adam SGD optimizer for all models. When dealing with larger datasets of $64 \times 64$, we reduce the batch size to $4$ and the representation dimensions to $30$. During experiments, for each model we conducted 10 runs on each dataset group by five different generation seeds and two different input shufflings. We saved one model snapshot every epoch and chose the best three according to their validation performance, and then computed the mean and standard deviation of their evaluation metrics on the test set. 

\textbf{Evaluation metrics}. We use Hits@1, Hits@5, Hits@10, the mean rank (MR), and the mean reciprocal rank (MRR) to evaluate these models. Hits@k means the proportion of test source-destination pairs for which the target destination is ranked in the top-k predictions, and thus Hits@1 is the prediction accuracy. Compared to Hits@k, MR and MRR can evaluate prediction results even when the target destination is ranked out of the top-k. MR provides a more intuitive sense about how many are ranked before the target on average, but often suffers from its instability susceptible to the worst example and becomes very large. MRR scores always range from $0.0$ to $1.0$. For MR lower score reflects better prediction, whereas for MRR higher score means better.

\begin{table}[t]
  \caption{\footnotesize{Comparison results on dataset groups \textit{\{LINE,SINE,LOCATION,HISTORY\}-SZ32-STP16-NDRP-STD0.5}. The marks in this table have the same meanings as Table \ref{tab:comparison-results-std0.2}}.}
  \label{tab:comparison-results-std0.5}
  \centering
  \begin{tabular}{l|cc|cc|cc|cc}
    \noalign{\hrule height 1.2pt}
    & \multicolumn{2}{c|}{\footnotesize{LINE}} & \multicolumn{2}{|c|}{\footnotesize{SINE}} & \multicolumn{2}{|c|}{\footnotesize{LOCATION}} & \multicolumn{2}{|c}{\footnotesize{HISTORY}} \\
    Model & \footnotesize{H@1(\%)} & \footnotesize{MRR} & \footnotesize{H@1(\%)} & \footnotesize{MRR} & \footnotesize{H@1(\%)} & \footnotesize{MRR} & \footnotesize{H@1(\%)} & \footnotesize{MRR} \\
    \noalign{\hrule height 1.2pt}
    \footnotesize{RW-Stationary} & \footnotesize{15.74}\scriptsize{$*$} & \footnotesize{0.3348}$\enspace$ & \footnotesize{9.07} & \footnotesize{0.2267}$\enspace$ & \footnotesize{19.40}\scriptsize{$*$} & \footnotesize{0.3860}\scriptsize{$*$} & \footnotesize{11.72}$\enspace$ & \footnotesize{0.2547}$\enspace$ \\
    \footnotesize{RW-Dynamic} & \footnotesize{15.48}$\enspace$ & \footnotesize{0.3429}\scriptsize{$*$} & \footnotesize{13.38}\scriptsize{$*$} & \footnotesize{0.2905}\scriptsize{$*$} & \footnotesize{17.91}$\enspace$ & \footnotesize{0.3722}$\enspace$ & \footnotesize{12.25}\scriptsize{$*$} & \footnotesize{0.2820}\scriptsize{$*$} \\
    \noalign{\hrule height 1.2pt}
    \footnotesize{FullGN} & \textcolor{mygray}{\textit{\footnotesize{14.61}}} & \textcolor{mygray}{\textit{\footnotesize{0.3355}}} & \textcolor{mygray}{\textit{\footnotesize{17.10}}} & \textcolor{mygray}{\textit{\footnotesize{0.3525}}} & \textcolor{mygray}{\textit{\footnotesize{16.09}}} & \textcolor{mygray}{\textit{\footnotesize{0.3476}}} & \textcolor{mygray}{\textit{\footnotesize{13.79}}} & \textcolor{mygray}{\textit{\footnotesize{0.3051}}} \\
    \cline{2-9}
    \footnotesize{FullGN-NoAct} & \footnotesize{15.50}$\;\:\,$ & \footnotesize{0.3410}$\;\:\,$ & \footnotesize{16.60}\tiny{$\checkmark$} & \footnotesize{0.3360}\tiny{$\checkmark$} & \footnotesize{18.83}$\;\:\,$ & \footnotesize{0.3816}$\;\:\,$ & \footnotesize{13.90}\tiny{$\checkmark$} & \footnotesize{0.3059}\tiny{$\checkmark$} \\
    \footnotesize{FullGN-Mul} & \footnotesize{16.21}\tiny{$\checkmark$} & \footnotesize{0.3498}\tiny{$\checkmark$} & \footnotesize{16.93}\tiny{$\checkmark$} & \footnotesize{0.3283}\tiny{$\checkmark$} & \footnotesize{18.50}$\;\:\,$ & \footnotesize{0.3787}$\;\:\,$ & \footnotesize{12.93}\tiny{$\checkmark$} & \footnotesize{0.2807}$\;\:\,$ \\
    \footnotesize{FullGN-MulMlp} & \textbf{\footnotesize{16.07}}\tiny{$\checkmark$} & \textbf{\footnotesize{0.3502}}\tiny{$\checkmark$} & \textbf{\footnotesize{17.31}}\tiny{$\checkmark$} & \textbf{\footnotesize{0.3389}}\tiny{$\checkmark$} & \textbf{\footnotesize{19.64}}\tiny{$\checkmark$} & \textbf{\footnotesize{0.3991}}\tiny{$\checkmark$} & \textbf{\footnotesize{14.89}}\tiny{$\checkmark$} & \textbf{\footnotesize{0.3145}}\tiny{$\checkmark$} \\
    \noalign{\hrule height 1.2pt}
    \footnotesize{GGNN} & \textcolor{mygray}{\textit{\footnotesize{14.53}}} & \textcolor{mygray}{\textit{\footnotesize{0.3344}}} & \textcolor{mygray}{\textit{\footnotesize{17.45}}} & \textcolor{mygray}{\textit{\footnotesize{0.3555}}} & \textcolor{mygray}{\textit{\footnotesize{17.11}}} & \textcolor{mygray}{\textit{\footnotesize{0.3689}}} & \textcolor{mygray}{\textit{\footnotesize{14.83}}} & \textcolor{mygray}{\textit{\footnotesize{0.3217}}} \\
    \cline{2-9}
    \footnotesize{GGNN-NoAct} & \footnotesize{15.58}$\;\:\,$ & \footnotesize{0.3415}$\;\:\,$ & \footnotesize{16.51}\tiny{$\checkmark$} & \footnotesize{0.3262}\tiny{$\checkmark$} & \textbf{\footnotesize{19.66}}\tiny{$\checkmark$} & \textbf{\footnotesize{0.3912}}\tiny{$\checkmark$} & \footnotesize{13.84}\tiny{$\checkmark$} & \footnotesize{0.2957}\tiny{$\checkmark$} \\
    \footnotesize{GGNN-Mul} & \footnotesize{15.79}\tiny{$\checkmark$} & \footnotesize{0.3448}\tiny{$\checkmark$} & \footnotesize{16.03}\tiny{$\checkmark$} & \footnotesize{0.3226}\tiny{$\checkmark$} & \footnotesize{17.84}$\;\:\,$ & \footnotesize{0.3723}$\;\:\,$ & \footnotesize{14.16}\tiny{$\checkmark$} & \footnotesize{0.2971}\tiny{$\checkmark$} \\
    \footnotesize{GGNN-MulMlp} & \textbf{\footnotesize{15.99}}\tiny{$\checkmark$} & \textbf{\footnotesize{0.3497}}\tiny{$\checkmark$} & \textbf{\footnotesize{17.31}}\tiny{$\checkmark$} & \textbf{\footnotesize{0.3370}}\tiny{$\checkmark$} & \footnotesize{19.39}$\;\:\,$ & \footnotesize{0.3911}\tiny{$\checkmark$} & \textbf{\footnotesize{14.80}}\tiny{$\checkmark$} & \textbf{\footnotesize{0.3053}}\tiny{$\checkmark$} \\
    \noalign{\hrule height 1.2pt}
    \footnotesize{GAT} & \textcolor{mygray}{\textit{\footnotesize{14.79}}} & \textcolor{mygray}{\textit{\footnotesize{0.3300}}} & \textcolor{mygray}{\textit{\footnotesize{16.48}}} & \textcolor{mygray}{\textit{\footnotesize{0.3338}}} & \textcolor{mygray}{\textit{\footnotesize{14.51}}} & \textcolor{mygray}{\textit{\footnotesize{0.3227}}} & \textcolor{mygray}{\textit{\footnotesize{10.80}}} & \textcolor{mygray}{\textit{\footnotesize{0.2538}}} \\
    \cline{2-9}
    \footnotesize{GAT-NoAct} & \footnotesize{15.83}\tiny{$\checkmark$} & \footnotesize{0.3414}$\;\:\,$ & \footnotesize{15.15}\tiny{$\checkmark$} & \footnotesize{0.3161}\tiny{$\checkmark$} & \footnotesize{17.82}$\;\:\,$ & \footnotesize{0.3702}$\;\:\,$ & \footnotesize{12.60}\tiny{$\checkmark$} & \footnotesize{0.2829}\tiny{$\checkmark$} \\
    \footnotesize{GAT-Mul} & \footnotesize{15.01}\tiny{$\checkmark$} & \footnotesize{0.3351}$\;\:\,$ & \footnotesize{14.85}\tiny{$\checkmark$} & \footnotesize{0.3070}\tiny{$\checkmark$} & \footnotesize{18.27}$\;\:\,$ & \footnotesize{0.3749}$\;\:\,$ & \footnotesize{13.49}\tiny{$\checkmark$} & \footnotesize{0.2885}\tiny{$\checkmark$} \\
    \footnotesize{GAT-MulMlp} & \textbf{\footnotesize{16.25}}\tiny{$\checkmark$} & \textbf{\footnotesize{0.3493}}\tiny{$\checkmark$} & \textbf{\footnotesize{16.45}}\tiny{$\checkmark$} & \textbf{\footnotesize{0.3292}}\tiny{$\checkmark$} & \textbf{\footnotesize{18.93}}$\;\:\,$ & \textbf{\footnotesize{0.3843}}$\;\:\,$ & \textbf{\footnotesize{13.75}}\tiny{$\checkmark$} & \textbf{\footnotesize{0.2933}}\tiny{$\checkmark$} \\
    \noalign{\hrule height 1.2pt}
  \end{tabular}
\end{table}

\begin{table}[t]
  \caption{\footnotesize{Comparison results on larger datasets of $64\times 64$ that are \textit{\{LINE,SINE,LOCATION,HISTORY\}-SZ64-STP32-NDRP-STD\{0.2,0.5\}}.}}
  \label{tab:comparison-results-64x64}
  \centering
  \begin{tabular}{ll|cc|cc|cc|cc}
    \noalign{\hrule height 1.2pt}
    & & \multicolumn{2}{c|}{\footnotesize{LINE}} & \multicolumn{2}{|c|}{\footnotesize{SINE}} & \multicolumn{2}{|c|}{\footnotesize{LOCATION}} & \multicolumn{2}{|c}{\footnotesize{HISTORY}} \\
    Std & Model & \footnotesize{H@1(\%)} & \footnotesize{MRR} & \footnotesize{H@1(\%)} & \footnotesize{MRR} & \footnotesize{H@1(\%)} & \footnotesize{MRR} & \footnotesize{H@1(\%)} & \footnotesize{MRR} \\
    \noalign{\hrule height 1.2pt}
    \multirow{2}{*}{\footnotesize{0.2}} & \footnotesize{GGNN} & \footnotesize{12.12} & \footnotesize{0.2887} & \footnotesize{32.11} & \footnotesize{0.4681} & \footnotesize{22.59} & \footnotesize{0.4335} & \footnotesize{9.99} & \footnotesize{0.2299} \\
    & \footnotesize{GGNN-MulMlp} & \textbf{\footnotesize{15.54}} & \textbf{\footnotesize{0.3324}} & \textbf{\footnotesize{38.21}} & \textbf{\footnotesize{0.5177}} & \textbf{\footnotesize{44.01}} & \textbf{\footnotesize{0.6286}} & \textbf{\footnotesize{24.54}} & \textbf{\footnotesize{0.3931}} \\
    \noalign{\hrule height 1.2pt}
    \multirow{2}{*}{\footnotesize{0.5}} & \footnotesize{GGNN} & \footnotesize{11.66} & \footnotesize{0.2814} & \footnotesize{10.47} & \footnotesize{0.2484} & \footnotesize{11.80} & \footnotesize{0.2784} & \footnotesize{7.65} & \footnotesize{0.2064} \\
    & \footnotesize{GGNN-MulMlp} & \textbf{\footnotesize{14.95}} & \textbf{\footnotesize{0.3231}} & \textbf{\footnotesize{16.68}} & \textbf{\footnotesize{0.3238}} & \textbf{\footnotesize{18.51}} & \textbf{\footnotesize{0.3778}} & \textbf{\footnotesize{13.47}} & \textbf{\footnotesize{0.2839}} \\
    \noalign{\hrule height 1.2pt}
  \end{tabular}
\end{table}

\subsection{Experimental Results}

\begin{figure}[t]
\centering
\includegraphics[width=0.85\textwidth]{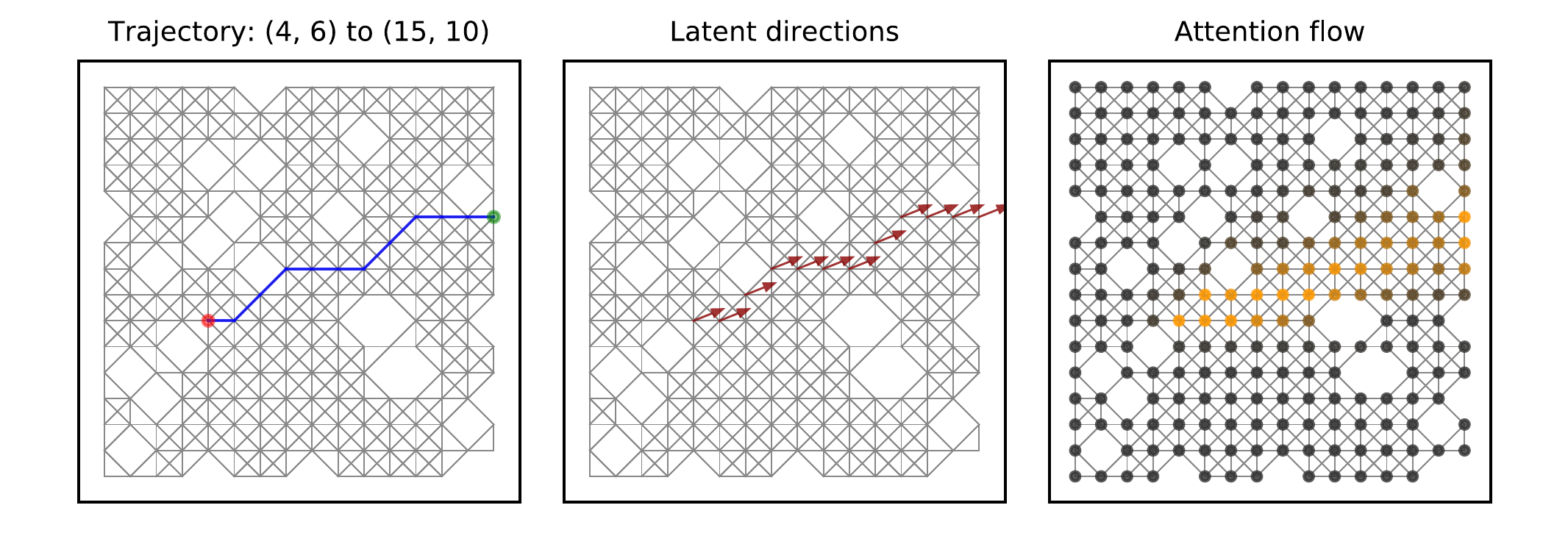}
\includegraphics[width=0.85\textwidth]{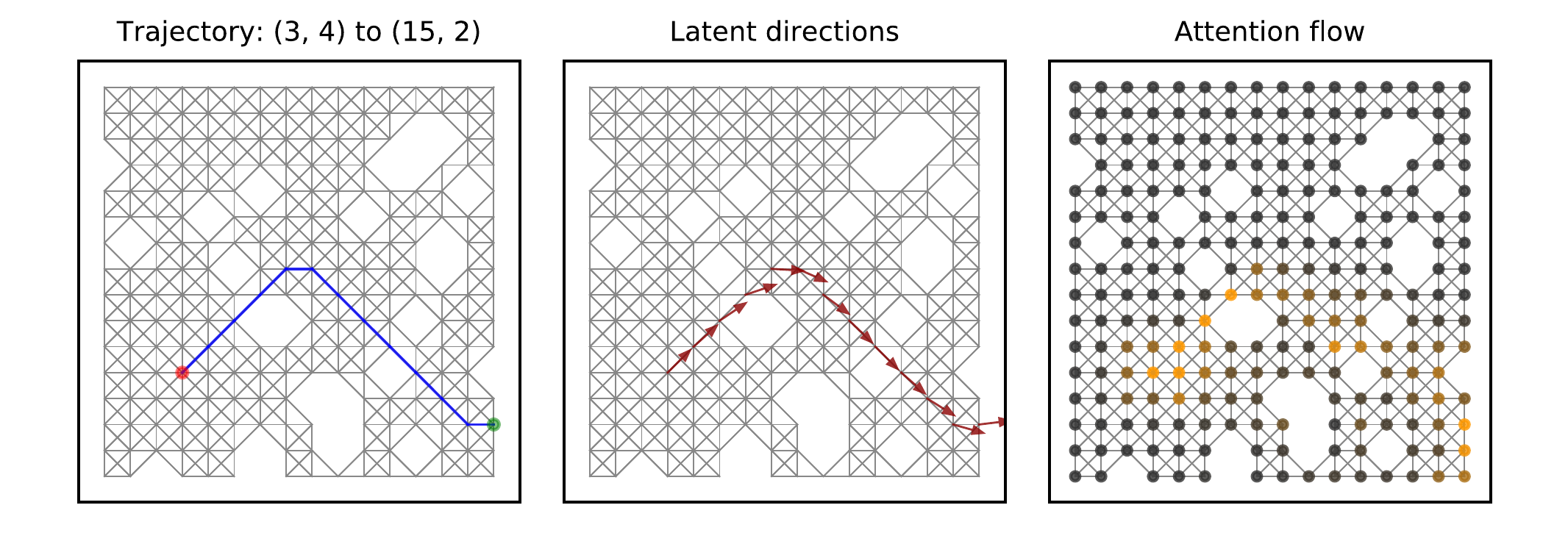}
\includegraphics[width=0.85\textwidth]{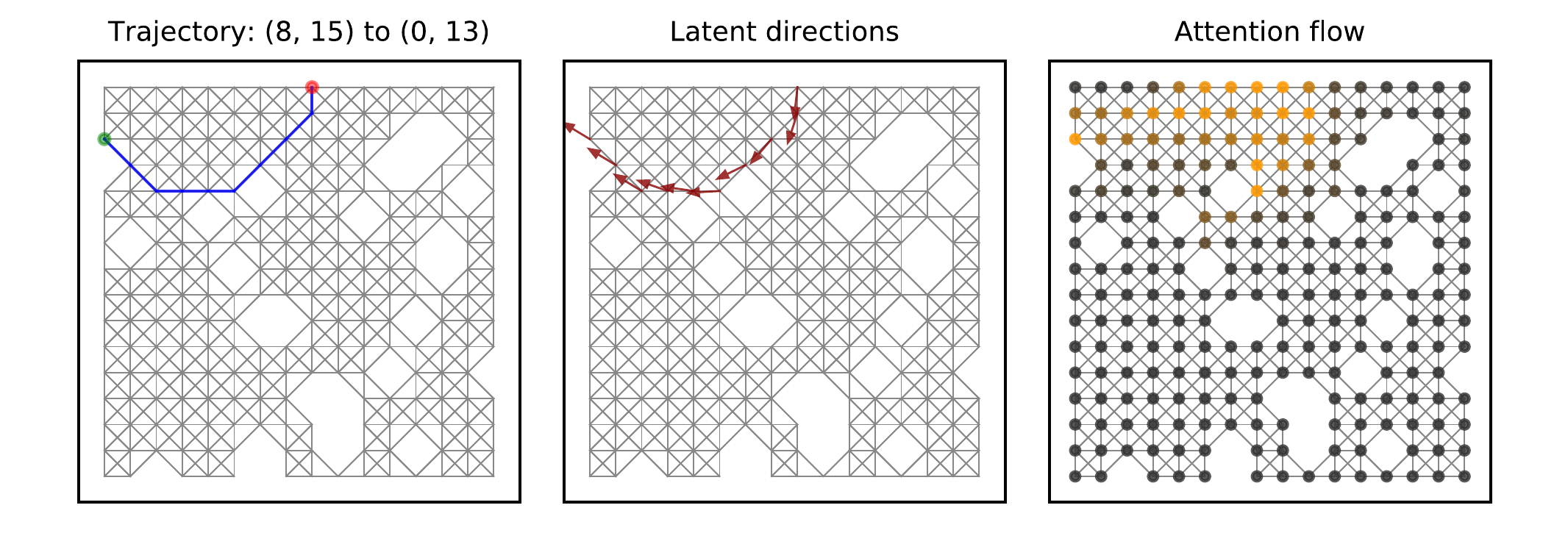}
\includegraphics[width=0.85\textwidth]{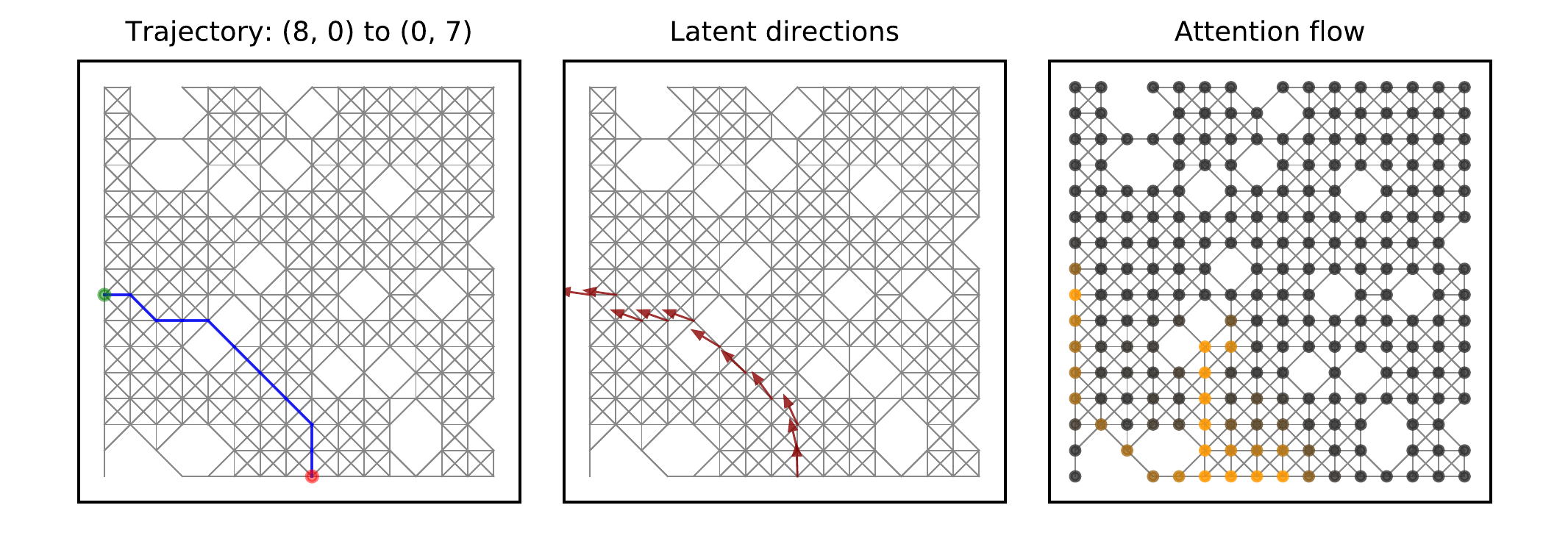}
\caption{\footnotesize{The true trajectory v.s. the latent directions in arrows v.s. the learned attention flow. The first row reflects a constant sloped direction, the second row time-dependent directions, the third row location-dependent directions, and the last row history-dependent directions. The drawn attention flow is based on the max aggregation of normalized attention distributions over the $T$ steps.}}
\label{fig:visual}
\vspace{-10pt}
\end{figure}

\textbf{Objectives of comparison}. We list our objectives of comparative evaluation from three aspects.
\begin{itemize}[nosep, wide=0pt, leftmargin=\dimexpr\labelwidth + 2\labelsep\relax]
    \item To test how well the explicit attention flow modeling can leverage rich context carried by message passing in graph networks, compared to the modeling purely based on random walks.
    \item To test whether the backward acting of attention flow on message passing is useful, and which way can be the most effective.
    \item To test whether the explicit attention flow can improve the prediction accuracy.
\end{itemize}

\textbf{Discussion on comparison results}. First, we compare the models that explicitly model attention flow, between the random walk-based and the graph network-based. From Table \ref{tab:comparison-results-std0.2} and \ref{tab:comparison-results-std0.5}, we see in most cases the models favored by graph networks surpass the random walk-based models, often by a large margin. Although there are exceptions that \textit{RW-Stationary} performs strong in the location-dependent cases, probably due to little context needed other than current location information, the best of the graph network-based models, such as \textit{FullGN-MulMlp}, can still beat it. Second, we compare the backward acting mechanism between no acting, the multiplying acting, and the non-linear acting. The non-linear acting after multiplying performs the best in almost all cases. What surprises us is that simply doing multiplying may degrade the performance, making it worse than no acting. How to design an effective backward acting mechanism is worth further study in future work. Last, we compare our remodeled graph networks with explicit attention flow against the regular graph networks. For the $32\times 32$ datasets, \textit{\{FullGN,GGNN,GAT\}-MulMlp} exceed their regular graph network counterparts respectively except for dataset groups \textit{\footnotesize{SINE-SZ32-STP16-*}}. For the larger $64\times 64$ datasets, we test \textit{GGNN} and \textit{GGNN-MulMlp}, and find that \textit{GGNN-MulMlp} performs significantly much better than \textit{GGNN} on every evaluation metric as shown in Table \ref{tab:comparison-results-64x64}.

\textbf{Discussion on visualization results}. We visualize the learned attention flow in a $16\times 16$ corrupted grid map, compared with the true trajectories and latent directions by taking one example for each direction setting as shown in Figure \ref{fig:visual}. At first glance, the drawn attention flow over the $T$ steps makes up a belt linking from the source node to the destination node, almost matching the true trajectories, especially as shown in the first and second rows in Figure \ref{fig:visual}. On closer inspection, we find that the attention flow might not necessarily follow the one-path pattern but instead branch to enlarge the exploring area that is more likely to contain a destination node especially near gaps, as shown in the last two rows.
\section{Conclusion}
In this paper, we introduce the attention flow mechanism to explicitly model the reasoning process on graphs, leveraging the rich context carried by message passing in graph networks. We treat this mechanism as a way to offer accurate predictions as well as clear interpretations. In addition, we study the backward acting of attention flow on information flow implemented by message passing, and show some interesting findings from experimental results. The interaction between the two flows, one favoring the fitting capacity and one offering the interpretability, may be worth further study in future work.

\bibliography{reference}{}
\bibliographystyle{unsrt}

\newpage

\setcounter{section}{0}
\begin{centering}
{\Large{\textbf{Appendix}}}
\end{centering}

\section{Experiments (cont'd)}

\begin{table}[h]
  \caption{Parameters of generating the datasets}
  \label{tab:params-gen}
  \centering
  \begin{tabular}{lcccccc}
    \toprule
    Dataset Group & $\vec{d}_{t,x,y}$ & $N$ & $T$ & $\sigma$ & $p_\mathrm{node\_drp}$ & $p_\mathrm{edge\_drp}$ \\
    \midrule
    \textit{\small LINE-SZ32-STP16-NDRP-STD0.2} & Line & 32 & 16 & 0.2 & 0.1 & 0.0 \\
    \textit{\small LINE-SZ32-STP16-NDRP-STD0.5} & Line & 32 & 16 & 0.5 & 0.1 & 0.0 \\
    \textit{\small SINE-SZ32-STP16-NDRP-STD0.2} & Sine & 32 & 16 & 0.2 & 0.1 & 0.0 \\
    \textit{\small SINE-SZ32-STP16-NDRP-STD0.5} & Sine & 32 & 16 & 0.5 & 0.1 & 0.0 \\
    \textit{\small LOCATION-SZ32-STP16-NDRP-STD0.2} & Location & 32 & 16 & 0.2 & 0.1 & 0.0 \\
    \textit{\small LOCATION-SZ32-STP16-NDRP-STD0.5} & Location & 32 & 16 & 0.5 & 0.1 & 0.0 \\
    \textit{\small HISTORY-SZ32-STP16-NDRP-STD0.2} & History & 32 & 16 & 0.2 & 0.1 & 0.0 \\
    \textit{\small HISTORY-SZ32-STP16-NDRP-STD0.5} & History & 32 & 16 & 0.5 & 0.1 & 0.0 \\
    \midrule
    \textit{\small LINE-SZ32-STP16-EDRP-STD0.2} & Line & 32 & 16 & 0.2 & 0.0 & 0.2 \\
    \textit{\small LINE-SZ32-STP16-EDRP-STD0.5} & Line & 32 & 16 & 0.5 & 0.0 & 0.2 \\
    \textit{\small SINE-SZ32-STP16-EDRP-STD0.2} & Sine & 32 & 16 & 0.2 & 0.0 & 0.2 \\
    \textit{\small SINE-SZ32-STP16-EDRP-STD0.5} & Sine & 32 & 16 & 0.5 & 0.0 & 0.2 \\
    \textit{\small LOCATION-SZ32-STP16-EDRP-STD0.2} & Location & 32 & 16 & 0.2 & 0.0 & 0.2 \\
    \textit{\small LOCATION-SZ32-STP16-EDRP-STD0.5} & Location & 32 & 16 & 0.5 & 0.0 & 0.2 \\
    \textit{\small HISTORY-SZ32-STP16-EDRP-STD0.2} & History & 32 & 16 & 0.2 & 0.0 & 0.2 \\
    \textit{\small HISTORY-SZ32-STP16-EDRP-STD0.5} & History & 32 & 16 & 0.5 & 0.0 & 0.2 \\
    \midrule
    \textit{\small LINE-SZ64-STP32-NDRP-STD0.2} & Line & 64 & 32 & 0.2 & 0.1 & 0.0 \\
    \textit{\small LINE-SZ64-STP32-NDRP-STD0.5} & Line & 64 & 32 & 0.5 & 0.1 & 0.0 \\
    \textit{\small SINE-SZ64-STP32-NDRP-STD0.2} & Sine & 64 & 32 & 0.2 & 0.1 & 0.0 \\
    \textit{\small SINE-SZ64-STP32-NDRP-STD0.5} & Sine & 64 & 32 & 0.5 & 0.1 & 0.0 \\
    \textit{\small LOCATION-SZ64-STP32-NDRP-STD0.2} & Location & 64 & 32 & 0.2 & 0.1 & 0.0 \\
    \textit{\small LOCATION-SZ64-STP32-NDRP-STD0.5} & Location & 64 & 32 & 0.5 & 0.1 & 0.0 \\
    \textit{\small HISTORY-SZ64-STP32-NDRP-STD0.2} & History & 64 & 32 & 0.2 & 0.1 & 0.0 \\
    \textit{\small HISTORY-SZ64-STP32-NDRP-STD0.5} & History & 64 & 32 & 0.5 & 0.1 & 0.0 \\
    \bottomrule
  \end{tabular}
\end{table}

\begin{table}[h]
  \caption{Dataset statistics \small{(All numbers represent average results. Note that dataset groups \textit{LINE-*-STD0.2} produce more trajectories per node than their counterparts, because the slop we choose makes the latent direction equally between two candidate edges, introducing more randomness to generate trajectories.)}}
  \label{tab:dataset-stats}
  \centering
  \begin{tabular}{lccccc}
    \toprule
    Dataset Group & \#Nodes & \#Edges & \#Trajs & \#Trajs-per-node & Traj-length \\
    \midrule
    \textit{\small LINE-SZ32-STP16-NDRP-STD0.2} & 921 & 6319 & 4829 & 5.2 & 9.1 \\
    \textit{\small LINE-SZ32-STP16-NDRP-STD0.5} & 921 & 6319 & 5029 & 5.5 & 9.0 \\
    \textit{\small SINE-SZ32-STP16-NDRP-STD0.2} & 921 & 6319 & 1555 & 1.7 & 9.0 \\
    \textit{\small SINE-SZ32-STP16-NDRP-STD0.5} & 921 & 6319 & 4380 & 4.8 & 9.4 \\
    \textit{\small LOCATION-SZ32-STP16-NDRP-STD0.2} & 921 & 6319 & 1440 & 1.6 & 7.9 \\
    \textit{\small LOCATION-SZ32-STP16-NDRP-STD0.5} & 921 & 6319 & 4098 & 4.4 & 8.8 \\
    \textit{\small HISTORY-SZ32-STP16-NDRP-STD0.2} & 921 & 6319 & 1541 & 1.7 & 8.5 \\
    \textit{\small HISTORY-SZ32-STP16-NDRP-STD0.5} & 921 & 6319 & 4418 & 4.8 & 9.2 \\
    \midrule
    \textit{\small LINE-SZ32-STP16-EDRP-STD0.2} & 1023 & 6248 & 4828 & 4.7 & 6.8 \\
    \textit{\small LINE-SZ32-STP16-EDRP-STD0.5} & 1023 & 6248 & 5051 & 4.9 & 6.7 \\
    \textit{\small SINE-SZ32-STP16-EDRP-STD0.2} & 1023 & 6248 & 1441 & 1.4 & 6.3 \\
    \textit{\small SINE-SZ32-STP16-EDRP-STD0.5} & 1023 & 6248 & 4238 & 4.1 & 6.8 \\
    \textit{\small LOCATION-SZ32-STP16-EDRP-STD0.2} & 1023 & 6248 & 1576 & 1.5 & 6.0 \\
    \textit{\small LOCATION-SZ32-STP16-EDRP-STD0.5} & 1023 & 6248 & 4327 & 4.2 & 6.6 \\
    \textit{\small HISTORY-SZ32-STP16-EDRP-STD0.} & 1023 & 6248 & 1586 & 1.5 & 6.0 \\
    \textit{\small HISTORY-SZ32-STP16-EDRP-STD0.} & 1023 & 6248 & 4441 & 4.3 & 6.6 \\
    \midrule
    \textit{\small LINE-SZ64-STP32-NDRP-STD0.2} & 3686 & 25891 & 21390 & 5.8 & 11.4 \\
    \textit{\small LINE-SZ64-STP32-NDRP-STD0.5} & 3686 & 25891 & 22106 & 6.0 & 11.1 \\
    \textit{\small SINE-SZ64-STP32-NDRP-STD0.2} & 3686 & 25891 & 6691 & 1.8 & 11.4 \\
    \textit{\small SINE-SZ64-STP32-NDRP-STD0.5} & 3686 & 25891 & 19233 & 5.2 & 11.7 \\
    \textit{\small LOCATION-SZ64-STP32-NDRP-STD0.2} & 3686 & 25891 & 6522 & 1.8 & 10.1 \\
    \textit{\small LOCATION-SZ64-STP32-NDRP-STD0.5} & 3686 & 25891 & 17482 & 4.7 & 10.8 \\
    \textit{\small HISTORY-SZ64-STP32-NDRP-STD0.2} & 3686 & 25891 & 6987 & 1.9 & 11.0 \\
    \textit{\small HISTORY-SZ64-STP32-NDRP-STD0.5} & 3686 & 25891 & 19303 & 5.2 & 11.7 \\
    \bottomrule
  \end{tabular}
\end{table}

\newpage

\subsection{Discussion about Results}
During the experiments, we found some results in our expectation as discussed in the model section, as well as some unexpected results that surprise us, probably worth further study in future work. Now we summarize them as follows:
\begin{itemize}[wide=10pt, leftmargin=\dimexpr\labelwidth + 2\labelsep\relax]
    \item \textbf{\textit{Backward acting of attention flow on information flow is useful, better than no acting in most cases.}}
    \begin{itemize}[wide=10pt, leftmargin=\dimexpr\labelwidth + 2\labelsep\relax]
        \item \textit{FullGN-\{Mul,MulMlp\}} both perform better than \textit{FullGN-NoAct} on Hits@1, Hits@5, Hits@10, MR and MRR for dataset groups \textit{\small LINE-SZ32-STP16-*}, on Hits@1, Hits@5, Hits@10 and MRR for dataset groups \textit{\small SINE-SZ32-STP16-NDRP-STD0.2, SINE-SZ32-STP16-EDRP-*, LOCATION-SZ32-STP16-EDRP-STD0.5}, on Hits@1 for dataset groups \textit{\small SINE-SZ32-STP16-NDRP-STD0.5, HISTORY-SZ32-STP16-*-STD0.2}. 
        \item \textit{GGNN-\{Mul,MulMlp\}} both perform better than \textit{GGNN-NoAct} on Hits@1, Hits@5, Hits@10, MR and MRR for dataset group \textit{\small LINE-SZ32-STP16-NDRP-STD0.2}, on Hits@1, MR and MRR for dataset group \textit{\small LINE-SZ32-STP16-NDRP-STD0.5}, on Hits@1 and MRR for dataset groups \textit{\small HISTORY-SZ32-STP16-NDRP-*}, on Hits@1 for dataset group \textit{\small SINE-SZ32-STP16-NDRP-STD0.5}.
        \item \textit{GAT-\{Mul,MulMlp\}} both perform better than \textit{GAT-NoAct} on Hits@1, Hits@5 and MRR for dataset groups \textit{\small SINE-SZ32-STP16-NDRP-STD0.2, LOCATION-SZ32-STP16-NDRP-STD0.5, HISTORY-SZ32-STP16-NDRP-STD0.5}.
    \end{itemize}
    \item \textbf{\textit{Simply applying multiplying to backward acting might cause degradation.}}
        \begin{itemize}[wide=10pt, leftmargin=\dimexpr\labelwidth + 2\labelsep\relax]
            \item \textit{\{FullGN,GGNN,GAT\}-Mul} perform poorly on dataset groups \textit{\small LOCATION-SZ32-STP16-*}.
            \item \textit{\{GGNN,GAT\}-Mul} perform poorly on dataset group \textit{\small SINE-SZ32-STP16-NDRP-STD0.5}.
            \item \textit{GGNN-Mul} performs poorly on dataset group \textit{\small SINE-SZ32-STP16-NDRP-STD0.2}.
        \end{itemize}
    \item \textbf{\textit{Non-linear backward acting after multiplying can work consistently well, always performing the best among the backward acting of NoAct, Mul and MulMlp, and even often the best among all the models.}}
        \begin{itemize}[wide=10pt, leftmargin=\dimexpr\labelwidth + 2\labelsep\relax]
            \item \textit{\{FullGN,GGNN,GAT\}-MulMlp} perform the best on Hits@1, Hits@5, MR and MR for dataset groups \textit{\small LINE-SZ32-STP16-*-STD0.2}, on Hits@1 and MRR for dataset groups \textit{\small LOCATION-SZ32-STP16-*-STD0.2}
            \item \textit{\{GGNN,GAT\}-MulMlp} perform the best on Hits@1, Hits@5 and MR for dataset groups \textit{\small LINE-SZ32-STP16-NDRP-STD0.5, HISTORY-SZ32-STP16-NDRP-STD0.5}, on MR for dataset group \textit{\small SINE-SZ32-STP16-NDRP-STD0.2}
            \item \textit{\{FullGN,GAT\}-MulMlp} perform the best on Hits@1, Hits@5, Hits@10 and MR for dataset group \textit{\small LOCATION-SZ32-STP16-NDRP-STD0.5}, on MR for dataset group \textit{\small SINE-SZ32-STP16-NDRP-STD0.2}, on MR and MRR for dataset group \textit{\small LINE-SZ32-STP16-NDRP-STD0.5}
            \item \textit{\{FullGN,GGNN\}-MulMlp} perform the best on Hits@1 and MRR for dataset group \textit{\small HISTORY-SZ32-STP16-NDRP-STD0.2}.
            \item \textit{FullGN-MulMlp} performs the best on Hits@1, Hits@5 and MRR  for dataset groups \textit{\small LINE-SZ32-STP16-EDRP-STD0.2, LOCATION-SZ32-STP16-EDRP-*}, on Hits@1 on dataset groups \textit{\small LINE-SZ32-STP16-EDRP-*, LOCATION-SZ32-STP16-EDRP-*, HISTORY-SZ32-STP16-EDRP-STD0.2, SINE-SZ32-STP16-EDRP-STD0.5}, on MRR for dataset groups \textit{\small LINE-SZ32-STP16-EDRP-STD0.5, SINE-SZ32-STP16-EDRP-STD0.5, LOCATION-SZ32-STP16-EDRP-STD0.5}
        \end{itemize}
    \item \textbf{\textit{Reasoning purely based on random walks with no message passing may be the worst in most cases, but in a few cases it can work surprisingly well.}}
         \begin{itemize}[wide=10pt, leftmargin=\dimexpr\labelwidth + 2\labelsep\relax]
            \item \textit{RW-Stationary} performs very poorly on dataset groups \textit{\small LINE-SZ32-STP16-NDRP-*, SINE-SZ32-STP16-NDRP-*, HISTORY-SZ32-STP16-NDRP-*} but surprisingly well on dataset groups \textit{\small LOCATION-SZ32-STP16-NDRP-*}. It is probably because under the location-dependent latent directions it requires little context except current location information to do the trajectory reasoning.
            \item When considering global context information, \textit{RW-Dynamic} works better than \textit{RW-Stationary} in the cases where \textit{RW-Stationary} performs poorly, but it obtains lower scores than \textit{RW-Stationary} on dataset groups \textit{\small LOCATION-SZ32-STP16-NDRP-*}, demonstrating again that little context is needed here.
        \end{itemize}
    \item \textbf{\textit{Regular graph networks work extremely well in the cases with the time-dependent latent directions, but they might not be suitable for other cases, such as the location-dependent and the history-dependent latent directions.}}
        \begin{itemize}[wide=10pt, leftmargin=\dimexpr\labelwidth + 2\labelsep\relax]
            \item \textit{FullGN, GGNN, GAT} obtain the highest scores and exceed the second by a large margin on dataset groups \textit{\small SINE-SZ32-STP16-*} but still get a large MR score.
            \item \textit{FullGN, GGNN, GAT} perform very poorly on dataset groups \textit{\small LOCATION-SZ32-STP16-*, HISTORY-SZ32-STP16-*} except \textit{GGNN} on \textit{\small HISTORY-SZ32-STP16-NDRP-STD0.5}.
        \end{itemize}
    \item \textbf{\textit{Models with attention flow taking non-linear backward acting might perform significantly better on a larger scale than those without}}.
        \begin{itemize}[wide=10pt, leftmargin=\dimexpr\labelwidth + 2\labelsep\relax]
            \item From the results on dataset groups \textit{\small \{LINE,SINE,LOCATION,HISTORY\}-SZ64-STP32-NDRP-STD\{0.2,0.5\}}, we can see \textit{GGNN-MulMlp} surpasses \textit{GGNN} by a large amount on every evaluation metric.
        \end{itemize}
\end{itemize}

\begin{table}[h]
  \caption{Comparison results on dataset group \textit{LINE-SZ32-STP16-NDRP-STD0.2}}
  \centering
  \begin{tabular}{lccccc}
    \toprule
    Model & Hits@1 ($\%$) & Hits@5 ($\%$) & Hits@10 ($\%$) & MR & MRR \\
    \midrule
    RW-Stationary & 15.80 \footnotesize{$\pm$0.56} & 56.95 \footnotesize{$\pm$1.96} & 78.25 \footnotesize{$\pm$2.31} & 9.857 \footnotesize{$\pm$1.167} & 0.3409 \footnotesize{$\pm$0.0098} \\
    RW-Dynamic & 16.64 \footnotesize{$\pm$0.65} & 59.65 \footnotesize{$\pm$1.89} & 82.13 \footnotesize{$\pm$1.77} & 7.061 \footnotesize{$\pm$0.594} & 0.3562 \footnotesize{$\pm$0.0084} \\
    \midrule
    FullGN & 15.13 \footnotesize{$\pm$0.74} & 59.75 \footnotesize{$\pm$2.24} & \underline{\textbf{83.83}} \footnotesize{$\pm$1.99} & 6.800 \footnotesize{$\pm$0.993} & 0.3451 \footnotesize{$\pm$0.0082} \\
    FullGN-NoAct & 16.65 \footnotesize{$\pm$0.96} & 59.35 \footnotesize{$\pm$2.08} & 82.50 \footnotesize{$\pm$1.82} & 6.628 \footnotesize{$\pm$0.392} & 0.3574 \footnotesize{$\pm$0.0102} \\
    FullGN-Mul & 16.69 \footnotesize{$\pm$0.86} & 61.20 \footnotesize{$\pm$1.66} & 83.58 \footnotesize{$\pm$2.24} & 6.399 \footnotesize{$\pm$0.609} & 0.3636 \footnotesize{$\pm$0.0108} \\
    FullGN-MulMlp & \textbf{16.99} \footnotesize{$\pm$0.91} & \underline{\textbf{61.59}} \footnotesize{$\pm$2.05} & 83.73 \footnotesize{$\pm$1.73} & \underline{\textbf{6.296}} \footnotesize{$\pm$0.615} & \textbf{0.3662} \footnotesize{$\pm$0.0109} \\
    \midrule
    GGNN & 15.49 \footnotesize{$\pm$0.98} & 60.30 \footnotesize{$\pm$1.80} & \textbf{83.60} \footnotesize{$\pm$1.96} & 6.605 \footnotesize{$\pm$0.580} & 0.3493 \footnotesize{$\pm$0.0100} \\
    GGNN-NoAct & 16.64 \footnotesize{$\pm$0.86} & 59.62 \footnotesize{$\pm$2.02} & 82.52 \footnotesize{$\pm$1.53} & 6.760 \footnotesize{$\pm$0.429} & 0.3570 \footnotesize{$\pm$0.0095} \\
    GGNN-Mul & 16.95 \footnotesize{$\pm$0.70} & 59.80 \footnotesize{$\pm$1.65} & 82.61 \footnotesize{$\pm$2.05} & 6.477 \footnotesize{$\pm$0.532} & 0.3610 \footnotesize{$\pm$0.0089} \\
    GGNN-MulMlp & \underline{\textbf{17.08}} \footnotesize{$\pm$0.77} & \textbf{61.45} \footnotesize{$\pm$1.82} & 83.54 \footnotesize{$\pm$1.89} & \textbf{6.316} \footnotesize{$\pm$0.684} & \underline{\textbf{0.3673}} \footnotesize{$\pm$0.0111} \\
    \midrule
    GAT & 16.01 \footnotesize{$\pm$1.06} & 58.86 \footnotesize{$\pm$1.82} & 80.78 \footnotesize{$\pm$1.91} & 18.306 \footnotesize{$\pm$16.927} & 0.3469 \footnotesize{$\pm$0.0121} \\
    GAT-NoAct & 16.02 \footnotesize{$\pm$0.65} & 59.42 \footnotesize{$\pm$1.95} & 82.29 \footnotesize{$\pm$1.70} & 6.707 \footnotesize{$\pm$0.397} & 0.3536 \footnotesize{$\pm$0.0091} \\
    GAT-Mul & 15.86 \footnotesize{$\pm$0.99} & 58.77 \footnotesize{$\pm$2.64} & 81.67 \footnotesize{$\pm$1.98} & 6.518 \footnotesize{$\pm$0.474} & 0.3501 \footnotesize{$\pm$0.0122} \\
    GAT-MulMlp & \textbf{17.07} \footnotesize{$\pm$0.79} & \textbf{60.60} \footnotesize{$\pm$1.70} & \textbf{83.22} \footnotesize{$\pm$2.18} & \textbf{6.488} \footnotesize{$\pm$0.670} & \textbf{0.3646} \footnotesize{$\pm$0.0100} \\
    \bottomrule
  \end{tabular}
\end{table}

\begin{table}[h]
  \caption{Comparison results on dataset group \textit{LINE-SZ32-STP16-NDRP-STD0.5}}
  \centering
  \begin{tabular}{lccccc}
    \toprule
    Model & Hits@1 ($\%$) & Hits@5 ($\%$) & Hits@10 ($\%$) & MR & MRR \\
    \midrule
    RW-Stationary & 15.74 \footnotesize{$\pm$0.54} & 55.28 \footnotesize{$\pm$1.15} & 76.40 \footnotesize{$\pm$1.24} & 10.087 \footnotesize{$\pm$0.610} & 0.3348 \footnotesize{$\pm$0.0040} \\
    RW-Dynamic & 15.48 \footnotesize{$\pm$0.82} & 57.07 \footnotesize{$\pm$1.30} & 79.19 \footnotesize{$\pm$1.04} & 7.735 \footnotesize{$\pm$0.345} & 0.3429 \footnotesize{$\pm$0.0071} \\
    \midrule
    FullGN & 14.61 \footnotesize{$\pm$0.90} & 57.97 \footnotesize{$\pm$1.20} & \underline{\textbf{80.90}} \footnotesize{$\pm$1.02} & 8.472 \footnotesize{$\pm$1.331} & 0.3355 \footnotesize{$\pm$0.0090} \\
    FullGN-NoAct & 15.50 \footnotesize{$\pm$0.48} & 57.09 \footnotesize{$\pm$1.31} & 79.55 \footnotesize{$\pm$1.10} & 7.761 \footnotesize{$\pm$0.416} & 0.3410 \footnotesize{$\pm$0.0064} \\
    FullGN-Mul & \textbf{16.21} \footnotesize{$\pm$0.35} & \underline{\textbf{58.18}} \footnotesize{$\pm$1.45} & 80.73 \footnotesize{$\pm$0.93} & 7.463 \footnotesize{$\pm$0.284} & 0.3498 \footnotesize{$\pm$0.0047} \\
    FullGN-MulMlp & 16.07 \footnotesize{$\pm$0.56} & 58.16 \footnotesize{$\pm$1.02} & 80.59 \footnotesize{$\pm$1.29} & \textbf{7.431} \footnotesize{$\pm$0.283} & \underline{\textbf{0.3502}} \footnotesize{$\pm$0.0046} \\
    \midrule
    GGNN & 14.53 \footnotesize{$\pm$0.72} & 57.33 \footnotesize{$\pm$1.01} & 80.34 \footnotesize{$\pm$1.25} & 7.856 \footnotesize{$\pm$0.669} & 0.3344 \footnotesize{$\pm$0.0073} \\
    GGNN-NoAct & 15.58 \footnotesize{$\pm$0.53} & 57.44 \footnotesize{$\pm$1.00} & 79.62 \footnotesize{$\pm$1.17} & 7.789 \footnotesize{$\pm$0.349} & 0.3415 \footnotesize{$\pm$0.0055} \\
    GGNN-Mul & 15.79 \footnotesize{$\pm$0.63} & 57.36 \footnotesize{$\pm$1.92} & 80.13 \footnotesize{$\pm$1.16} & \textbf{7.387} \footnotesize{$\pm$0.303} & 0.3448 \footnotesize{$\pm$0.0060} \\
    GGNN-MulMlp & \textbf{15.99} \footnotesize{$\pm$0.59} & \textbf{58.17} \footnotesize{$\pm$1.26} & \textbf{80.79} \footnotesize{$\pm$0.98} & 7.391 \footnotesize{$\pm$0.325} & \textbf{0.3497} \footnotesize{$\pm$0.0052} \\
    \midrule
    GAT & 14.79 \footnotesize{$\pm$1.17} & 56.51 \footnotesize{$\pm$2.26} & 79.20 \footnotesize{$\pm$1.79} & 16.323 \footnotesize{$\pm$12.684} & 0.3300 \footnotesize{$\pm$0.0147} \\
    GAT-NoAct & 15.83 \footnotesize{$\pm$0.39} & 56.95 \footnotesize{$\pm$1.25} & 79.28 \footnotesize{$\pm$1.31} & 7.631 \footnotesize{$\pm$0.309} & 0.3414 \footnotesize{$\pm$0.0047} \\
    GAT-Mul & 15.01 \footnotesize{$\pm$0.84} & 55.95 \footnotesize{$\pm$1.22} & 78.55 \footnotesize{$\pm$1.62} & 7.793 \footnotesize{$\pm$0.530} & 0.3351 \footnotesize{$\pm$0.0071} \\
    GAT-MulMlp & \underline{\textbf{16.25}} \footnotesize{$\pm$0.57} & \textbf{57.61} \footnotesize{$\pm$1.35} & \textbf{80.60} \footnotesize{$\pm$0.80} & \underline{\textbf{7.292}} \footnotesize{$\pm$0.190} & \textbf{0.3493} \footnotesize{$\pm$0.0056} \\
    \bottomrule
  \end{tabular}
\end{table}

\begin{table}[h]
  \caption{Comparison results on dataset group \textit{SINE-SZ32-STP16-NDRP-STD0.2}}
  \centering
  \begin{tabular}{lccccc}
    \toprule
    Model & Hits@1 ($\%$) & Hits@5 ($\%$) & Hits@10 ($\%$) & MR & MRR \\
    \midrule
    RW-Stationary & 8.56 \footnotesize{$\pm$1.76} & 35.81 \footnotesize{$\pm$1.81} & 51.33 \footnotesize{$\pm$2.47} & 23.883 \footnotesize{$\pm$2.757} & 0.2177 \footnotesize{$\pm$0.0103} \\
    RW-Dynamic & 19.15 \footnotesize{$\pm$3.02} & 50.69 \footnotesize{$\pm$4.20} & 67.13 \footnotesize{$\pm$3.58} & 13.554 \footnotesize{$\pm$2.560} & 0.3418 \footnotesize{$\pm$0.0266} \\
    \midrule
    FullGN & \underline{\textbf{51.71}} \footnotesize{$\pm$3.46} & \underline{\textbf{87.46}} \footnotesize{$\pm$3.93} & \textbf{93.20} \footnotesize{$\pm$4.78} & 17.588 \footnotesize{$\pm$19.805} & \underline{\textbf{0.6665}} \footnotesize{$\pm$0.0348} \\
    FullGN-NoAct & 30.10 \footnotesize{$\pm$6.69} & 61.67 \footnotesize{$\pm$7.22} & 76.88 \footnotesize{$\pm$5.48} & \underline{\textbf{9.329}} \footnotesize{$\pm$2.224} & 0.4476 \footnotesize{$\pm$0.0597} \\
    FullGN-Mul & 37.49 \footnotesize{$\pm$1.97} & 63.60 \footnotesize{$\pm$2.72} & 76.91 \footnotesize{$\pm$2.31} & 9.800 \footnotesize{$\pm$1.077} & 0.4915 \footnotesize{$\pm$0.0212} \\
    FullGN-MulMlp & 39.91 \footnotesize{$\pm$2.80} & 66.15 \footnotesize{$\pm$3.01} & 78.19 \footnotesize{$\pm$2.33} & 9.377 \footnotesize{$\pm$0.824} & 0.5195 \footnotesize{$\pm$0.0242} \\
    \midrule
    GGNN & \textbf{51.02} \footnotesize{$\pm$2.10} & \textbf{87.15} \footnotesize{$\pm$3.44} & \underline{\textbf{93.51}} \footnotesize{$\pm$2.92} & 12.796 \footnotesize{$\pm$18.578} & \textbf{0.6611} \footnotesize{$\pm$0.0208} \\
    GGNN-NoAct & 25.14 \footnotesize{$\pm$6.53} & 54.30 \footnotesize{$\pm$6.68} & 69.91 \footnotesize{$\pm$4.57} & 12.509 \footnotesize{$\pm$2.212} & 0.3918 \footnotesize{$\pm$0.0604} \\
    GGNN-Mul & 23.62 \footnotesize{$\pm$2.30} & 53.62 \footnotesize{$\pm$2.64} & 67.37 \footnotesize{$\pm$3.12} & 15.002 \footnotesize{$\pm$1.354} & 0.3776 \footnotesize{$\pm$0.0168} \\
    GGNN-MulMlp & 34.75 \footnotesize{$\pm$2.93} & 60.76 \footnotesize{$\pm$4.26} & 73.72 \footnotesize{$\pm$3.33} & \textbf{12.392} \footnotesize{$\pm$1.525} & 0.4699 \footnotesize{$\pm$0.0264} \\
    \midrule
    GAT & \textbf{43.19} \footnotesize{$\pm$3.24} & \textbf{73.32} \footnotesize{$\pm$3.67} & \textbf{79.30} \footnotesize{$\pm$3.32} & 62.170 \footnotesize{$\pm$12.886} & \textbf{0.5566} \footnotesize{$\pm$0.0294} \\
    GAT-NoAct & 15.77 \footnotesize{$\pm$3.97} & 50.17 \footnotesize{$\pm$4.34} & 67.36 \footnotesize{$\pm$3.93} & 13.988 \footnotesize{$\pm$1.933} & 0.3221 \footnotesize{$\pm$0.0336} \\
    GAT-Mul & 20.14 \footnotesize{$\pm$2.10} & 50.70 \footnotesize{$\pm$2.94} & 66.65 \footnotesize{$\pm$2.81} & 15.725 \footnotesize{$\pm$1.922} & 0.3429 \footnotesize{$\pm$0.0193} \\
    GAT-MulMlp & 30.64 \footnotesize{$\pm$2.37} & 58.59 \footnotesize{$\pm$3.91} & 73.46 \footnotesize{$\pm$3.34} & \textbf{12.022} \footnotesize{$\pm$1.639} & 0.4390 \footnotesize{$\pm$0.0216} \\
    \bottomrule
  \end{tabular}
\end{table}

\begin{table}[h]
  \caption{Comparison results on dataset group \textit{SINE-SZ32-STP16-NDRP-STD0.5}}
  \centering
  \begin{tabular}{lccccc}
    \toprule
    Model & Hits@1 ($\%$) & Hits@5 ($\%$) & Hits@10 ($\%$) & MR & MRR \\
    \midrule
    RW-Stationary & 9.07 \footnotesize{$\pm$0.60} & 34.71 \footnotesize{$\pm$1.38} & 50.39 \footnotesize{$\pm$3.15} & 19.937 \footnotesize{$\pm$1.535} & 0.2267 \footnotesize{$\pm$0.0090} \\
    RW-Dynamic & 13.38 \footnotesize{$\pm$1.06} & 46.26 \footnotesize{$\pm$1.98} & 64.54 \footnotesize{$\pm$2.05} & 12.380 \footnotesize{$\pm$1.032} & 0.2905 \footnotesize{$\pm$0.0103} \\
    \midrule
    FullGN & 17.10 \footnotesize{$\pm$0.67} & \underline{\textbf{57.91}} \footnotesize{$\pm$1.90} & \textbf{76.37} \footnotesize{$\pm$1.58} & \underline{\textbf{9.517}} \footnotesize{$\pm$1.175} & \textbf{0.3525} \footnotesize{$\pm$0.0069} \\
    FullGN-NoAct & 16.60 \footnotesize{$\pm$1.13} & 52.81 \footnotesize{$\pm$1.95} & 71.91 \footnotesize{$\pm$1.58} & 9.542 \footnotesize{$\pm$0.790} & 0.3360 \footnotesize{$\pm$0.0106} \\
    FullGN-Mul & 16.93 \footnotesize{$\pm$0.99} & 50.41 \footnotesize{$\pm$1.98} & 67.74 \footnotesize{$\pm$1.89} & 12.163 \footnotesize{$\pm$0.526} & 0.3283 \footnotesize{$\pm$0.0133} \\
    FullGN-MulMlp & \textbf{17.31} \footnotesize{$\pm$0.79} & 52.74 \footnotesize{$\pm$1.68} & 70.69 \footnotesize{$\pm$2.67} & 11.328 \footnotesize{$\pm$1.042} & 0.3389 \footnotesize{$\pm$0.0072} \\
    \midrule
    GGNN & \underline{\textbf{17.45}} \footnotesize{$\pm$1.03} & \textbf{57.80} \footnotesize{$\pm$1.59} & \underline{\textbf{76.72}} \footnotesize{$\pm$2.14} & \textbf{10.011} \footnotesize{$\pm$1.730} & \underline{\textbf{0.3555}} \footnotesize{$\pm$0.0098} \\
    GGNN-NoAct & 16.51 \footnotesize{$\pm$1.51} & 50.63 \footnotesize{$\pm$3.06} & 69.49 \footnotesize{$\pm$2.48} & 10.611 \footnotesize{$\pm$1.031} & 0.3262 \footnotesize{$\pm$0.0163} \\
    GGNN-Mul & 16.03 \footnotesize{$\pm$1.21} & 50.61 \footnotesize{$\pm$1.80} & 69.17 \footnotesize{$\pm$2.34} & 11.253 \footnotesize{$\pm$0.514} & 0.3226 \footnotesize{$\pm$0.0156} \\
    GGNN-MulMlp & 17.31 \footnotesize{$\pm$1.55} & 53.07 \footnotesize{$\pm$2.75} & 70.65 \footnotesize{$\pm$1.08} & 11.500 \footnotesize{$\pm$0.705} & 0.3370 \footnotesize{$\pm$0.0177} \\
    \midrule
    GAT & \textbf{16.48} \footnotesize{$\pm$0.91} & \textbf{53.47} \footnotesize{$\pm$1.60} & \textbf{71.89} \footnotesize{$\pm$1.70} & 21.869 \footnotesize{$\pm$12.869} & \textbf{0.3338} \footnotesize{$\pm$0.0070} \\
    GAT-NoAct & 15.15 \footnotesize{$\pm$1.34} & 49.65 \footnotesize{$\pm$3.17} & 68.24 \footnotesize{$\pm$2.86} & \textbf{11.026} \footnotesize{$\pm$1.152} & 0.3161 \footnotesize{$\pm$0.0153} \\
    GAT-Mul & 14.85 \footnotesize{$\pm$0.90} & 48.34 \footnotesize{$\pm$2.27} & 67.67 \footnotesize{$\pm$2.24} & 11.681 \footnotesize{$\pm$0.363} & 0.3070 \footnotesize{$\pm$0.0097} \\
    GAT-MulMlp & 16.45 \footnotesize{$\pm$1.18} & 51.23 \footnotesize{$\pm$2.81} & 68.86 \footnotesize{$\pm$2.24} & 11.701 \footnotesize{$\pm$0.526} & 0.3292 \footnotesize{$\pm$0.0169} \\
    \bottomrule
  \end{tabular}
\end{table}

\begin{table}[h]
  \caption{Comparison results on dataset group \textit{LOCATION-SZ32-STP16-NDRP-STD0.2}}
  \centering
  \begin{tabular}{lccccc}
    \toprule
    Model & Hits@1 ($\%$) & Hits@5 ($\%$) & Hits@10 ($\%$) & MR & MRR \\
    \midrule
    RW-Stationary & 50.06 \footnotesize{$\pm$3.23} & 84.70 \footnotesize{$\pm$5.23} & 91.99 \footnotesize{$\pm$3.76} & 5.997 \footnotesize{$\pm$2.266} & 0.6625 \footnotesize{$\pm$0.0403} \\
    RW-Dynamic & 45.94 \footnotesize{$\pm$6.63} & 85.81 \footnotesize{$\pm$5.03} & 93.05 \footnotesize{$\pm$2.62} & 6.036 \footnotesize{$\pm$2.205} & 0.6320 \footnotesize{$\pm$0.0544} \\
    \midrule
    FullGN & 25.67 \footnotesize{$\pm$8.66} & 69.33 \footnotesize{$\pm$10.43} & 80.63 \footnotesize{$\pm$7.18} & 18.589 \footnotesize{$\pm$9.211} & 0.4393 \footnotesize{$\pm$0.0884} \\
    FullGN-NoAct & 46.47 \footnotesize{$\pm$3.86} & 84.30 \footnotesize{$\pm$3.88} & \textbf{92.22} \footnotesize{$\pm$2.64} & \textbf{5.797} \footnotesize{$\pm$2.316} & 0.6337 \footnotesize{$\pm$0.0346} \\
    FullGN-Mul & 43.44 \footnotesize{$\pm$6.02} & 82.88 \footnotesize{$\pm$5.66} & 89.49 \footnotesize{$\pm$4.43} & 7.395 \footnotesize{$\pm$2.546} & 0.6029 \footnotesize{$\pm$0.0511} \\
    FullGN-MulMlp & \underline{\textbf{50.93}} \footnotesize{$\pm$3.21} & \textbf{85.46} \footnotesize{$\pm$4.93} & 91.03 \footnotesize{$\pm$3.72} & 7.649 \footnotesize{$\pm$3.921} & \textbf{0.6598} \footnotesize{$\pm$0.0362} \\
    \midrule
    GGNN & 29.20 \footnotesize{$\pm$7.85} & 70.59 \footnotesize{$\pm$8.83} & 79.83 \footnotesize{$\pm$5.75} & 16.471 \footnotesize{$\pm$6.033} & 0.4699 \footnotesize{$\pm$0.0742} \\
    GGNN-NoAct & 49.50 \footnotesize{$\pm$3.91} & \underline{\textbf{87.84}} \footnotesize{$\pm$4.27} & \underline{\textbf{93.48}} \footnotesize{$\pm$3.09} & \textbf{5.850} \footnotesize{$\pm$2.498} & 0.6621 \footnotesize{$\pm$0.0383} \\
    GGNN-Mul & 45.22 \footnotesize{$\pm$4.67} & 83.48 \footnotesize{$\pm$5.20} & 90.46 \footnotesize{$\pm$3.57} & 8.098 \footnotesize{$\pm$3.410} & 0.6185 \footnotesize{$\pm$0.0437} \\
    GGNN-MulMlp & \textbf{50.28} \footnotesize{$\pm$3.36} & 86.41 \footnotesize{$\pm$4.16} & 91.55 \footnotesize{$\pm$3.27} & 6.634 \footnotesize{$\pm$2.951} & \underline{\textbf{0.6637}} \footnotesize{$\pm$0.0342} \\
    \midrule
    GAT & 18.18 \footnotesize{$\pm$4.96} & 59.84 \footnotesize{$\pm$6.92} & 73.31 \footnotesize{$\pm$5.37} & 21.609 \footnotesize{$\pm$3.249} & 0.3583 \footnotesize{$\pm$0.0579} \\
    GAT-NoAct & 46.10 \footnotesize{$\pm$4.34} & 85.67 \footnotesize{$\pm$4.78} & 92.54 \footnotesize{$\pm$3.61} & \underline{\textbf{5.332}} \footnotesize{$\pm$1.829} & 0.6356 \footnotesize{$\pm$0.0400} \\
    GAT-Mul & 45.83 \footnotesize{$\pm$2.74} & 82.99 \footnotesize{$\pm$3.29} & 89.72 \footnotesize{$\pm$2.19} & 7.059 \footnotesize{$\pm$1.654} & 0.6208 \footnotesize{$\pm$0.0232} \\
    GAT-MulMlp & \textbf{47.52} \footnotesize{$\pm$7.39} & \textbf{85.68} \footnotesize{$\pm$4.04} & \textbf{92.97} \footnotesize{$\pm$2.80} & 6.276 \footnotesize{$\pm$2.278} & \textbf{0.6371} \footnotesize{$\pm$0.0597} \\
    \bottomrule
  \end{tabular}
\end{table}

\begin{table}[h]
  \caption{Comparison results on dataset group \textit{LOCATION-SZ32-STP16-NDRP-STD0.5}}
  \centering
  \begin{tabular}{lccccc}
    \toprule
    Model & Hits@1 ($\%$) & Hits@5 ($\%$) & Hits@10 ($\%$) & MR & MRR \\
    \midrule
    RW-Stationary & 19.40 \footnotesize{$\pm$1.03} & 64.37 \footnotesize{$\pm$1.68} & 80.30 \footnotesize{$\pm$1.63} & 9.569 \footnotesize{$\pm$1.020} & 0.3860 \footnotesize{$\pm$0.0113} \\
    RW-Dynamic & 17.91 \footnotesize{$\pm$0.75} & 62.47 \footnotesize{$\pm$1.60} & 80.55 \footnotesize{$\pm$1.87} & 9.379 \footnotesize{$\pm$1.392} & 0.3722 \footnotesize{$\pm$0.0101} \\
    \midrule
    FullGN & 16.09 \footnotesize{$\pm$1.51} & 58.99 \footnotesize{$\pm$4.85} & 78.30 \footnotesize{$\pm$4.49} & 16.608 \footnotesize{$\pm$6.832} & 0.3476 \footnotesize{$\pm$0.0256} \\
    FullGN-NoAct & 18.83 \footnotesize{$\pm$1.17} & 63.27 \footnotesize{$\pm$1.71} & 82.63 \footnotesize{$\pm$1.61} & \textbf{8.140} \footnotesize{$\pm$0.986} & 0.3816 \footnotesize{$\pm$0.0111} \\
    FullGN-Mul & 18.50 \footnotesize{$\pm$1.10} & 62.79 \footnotesize{$\pm$1.96} & 81.34 \footnotesize{$\pm$1.55} & 8.928 \footnotesize{$\pm$1.070} & 0.3787 \footnotesize{$\pm$0.0099} \\
    FullGN-MulMlp & \textbf{19.64} \footnotesize{$\pm$1.02} & \underline{\textbf{66.33}} \footnotesize{$\pm$1.75} & \underline{\textbf{83.87}} \footnotesize{$\pm$1.57} & 8.235 \footnotesize{$\pm$1.025} & \underline{\textbf{0.3991}} \footnotesize{$\pm$0.0100} \\
    \midrule
    GGNN & 17.11 \footnotesize{$\pm$1.07} & 63.48 \footnotesize{$\pm$2.29} & 81.83 \footnotesize{$\pm$1.95} & 14.964 \footnotesize{$\pm$3.551} & 0.3689 \footnotesize{$\pm$0.0100} \\
    GGNN-NoAct & \underline{\textbf{19.66}} \footnotesize{$\pm$0.81} & 64.74 \footnotesize{$\pm$1.20} & 82.97 \footnotesize{$\pm$0.98} & \underline{\textbf{8.118}} \footnotesize{$\pm$0.762} & \textbf{0.3912} \footnotesize{$\pm$0.0086} \\
    GGNN-Mul & 17.84 \footnotesize{$\pm$1.11} & 62.36 \footnotesize{$\pm$1.50} & 81.60 \footnotesize{$\pm$2.31} & 9.220 \footnotesize{$\pm$1.276} & 0.3723 \footnotesize{$\pm$0.0114} \\
    GGNN-MulMlp & 19.39 \footnotesize{$\pm$0.60} & \textbf{65.81} \footnotesize{$\pm$1.65} & \textbf{83.57} \footnotesize{$\pm$1.40} & 8.375 \footnotesize{$\pm$0.746} & 0.3911 \footnotesize{$\pm$0.0061} \\
    \midrule
    GAT & 14.51 \footnotesize{$\pm$1.17} & 55.19 \footnotesize{$\pm$2.14} & 72.87 \footnotesize{$\pm$2.42} & 28.074 \footnotesize{$\pm$4.756} & 0.3227 \footnotesize{$\pm$0.0110} \\
    GAT-NoAct & 17.82 \footnotesize{$\pm$0.96} & 61.62 \footnotesize{$\pm$1.47} & 80.28 \footnotesize{$\pm$2.40} & 9.147 \footnotesize{$\pm$1.037} & 0.3702 \footnotesize{$\pm$0.0082} \\
    GAT-Mul & 18.27 \footnotesize{$\pm$0.73} & 62.42 \footnotesize{$\pm$2.34} & 80.82 \footnotesize{$\pm$1.26} & 9.260 \footnotesize{$\pm$1.066} & 0.3749 \footnotesize{$\pm$0.0093} \\
    GAT-MulMlp & \textbf{18.93} \footnotesize{$\pm$1.18} & \textbf{64.53} \footnotesize{$\pm$1.90} & \textbf{83.06} \footnotesize{$\pm$1.92} & \textbf{8.283} \footnotesize{$\pm$0.988} & \textbf{0.3843} \footnotesize{$\pm$0.0114} \\
    \bottomrule
  \end{tabular}
\end{table}

\begin{table}[h]
  \caption{Comparison results on dataset group \textit{HISTORY-SZ32-STP16-NDRP-STD0.2}}
  \centering
  \begin{tabular}{lccccc}
    \toprule
    Model & Hits@1 ($\%$) & Hits@5 ($\%$) & Hits@10 ($\%$) & MR & MRR \\
    \midrule
    RW-Stationary & 16.44 \footnotesize{$\pm$3.08} & 47.22 \footnotesize{$\pm$3.67} & 59.40 \footnotesize{$\pm$4.04} & 34.573 \footnotesize{$\pm$4.230} & 0.3215 \footnotesize{$\pm$0.0319} \\
    RW-Dynamic & 20.41 \footnotesize{$\pm$2.07} & 55.07 \footnotesize{$\pm$2.87} & 68.41 \footnotesize{$\pm$2.77} & 24.968 \footnotesize{$\pm$3.194} & 0.3656 \footnotesize{$\pm$0.0206} \\
    \midrule
    FullGN & 16.61 \footnotesize{$\pm$4.79} & 48.68 \footnotesize{$\pm$12.52} & 59.51 \footnotesize{$\pm$14.14} & 63.999 \footnotesize{$\pm$22.762} & 0.3095 \footnotesize{$\pm$0.0765} \\
    FullGN-NoAct & 20.80 \footnotesize{$\pm$2.37} & \textbf{57.16} \footnotesize{$\pm$3.94} & \underline{\textbf{70.07}} \footnotesize{$\pm$4.00} & \underline{\textbf{21.744}} \footnotesize{$\pm$3.576} & 0.3729 \footnotesize{$\pm$0.0243} \\
    FullGN-Mul & 21.35 \footnotesize{$\pm$3.31} & 52.83 \footnotesize{$\pm$3.56} & 64.58 \footnotesize{$\pm$3.05} & 30.538 \footnotesize{$\pm$3.866} & 0.3618 \footnotesize{$\pm$0.0286} \\
    FullGN-MulMlp & \textbf{23.94} \footnotesize{$\pm$1.75} & 54.98 \footnotesize{$\pm$3.86} & 65.76 \footnotesize{$\pm$3.01} & 30.251 \footnotesize{$\pm$5.933} & \textbf{0.3850} \footnotesize{$\pm$0.0230} \\
    \midrule
    GGNN & 22.56 \footnotesize{$\pm$2.43} & 53.52 \footnotesize{$\pm$3.10} & 62.88 \footnotesize{$\pm$2.97} & 70.509 \footnotesize{$\pm$11.580} & 0.3677 \footnotesize{$\pm$0.0179} \\
    GGNN-NoAct & 21.69 \footnotesize{$\pm$2.26} & \underline{\textbf{57.50}} \footnotesize{$\pm$4.18} & \textbf{69.54} \footnotesize{$\pm$3.70} & \textbf{24.676} \footnotesize{$\pm$2.777} & 0.3818 \footnotesize{$\pm$0.0186} \\
    GGNN-Mul & 23.81 \footnotesize{$\pm$2.21} & 54.66 \footnotesize{$\pm$2.18} & 65.62 \footnotesize{$\pm$2.62} & 29.538 \footnotesize{$\pm$3.694} & 0.3824 \footnotesize{$\pm$0.0175} \\
    GGNN-MulMlp & \underline{\textbf{26.06}} \footnotesize{$\pm$1.51} & 56.04 \footnotesize{$\pm$2.94} & 66.18 \footnotesize{$\pm$3.54} & 32.089 \footnotesize{$\pm$5.656} & \underline{\textbf{0.4001}} \footnotesize{$\pm$0.0173} \\
    \midrule
    GAT & 12.11 \footnotesize{$\pm$1.71} & 36.94 \footnotesize{$\pm$3.05} & 47.97 \footnotesize{$\pm$4.17} & 94.705 \footnotesize{$\pm$8.713} & 0.2333 \footnotesize{$\pm$0.0184} \\
    GAT-NoAct & \textbf{23.17} \footnotesize{$\pm$2.02} & 56.06 \footnotesize{$\pm$3.09} & \textbf{68.00} \footnotesize{$\pm$2.89} & \textbf{27.490} \footnotesize{$\pm$3.953} & \textbf{0.3818} \footnotesize{$\pm$0.0186} \\
    GAT-Mul & 22.70 \footnotesize{$\pm$2.58} & \textbf{56.12} \footnotesize{$\pm$3.22} & 67.75 \footnotesize{$\pm$3.05} & 30.113 \footnotesize{$\pm$3.024} & 0.3762 \footnotesize{$\pm$0.0188} \\
    GAT-MulMlp & 20.71 \footnotesize{$\pm$2.98} & 54.86 \footnotesize{$\pm$2.76} & 66.01 \footnotesize{$\pm$3.23} & 29.533 \footnotesize{$\pm$5.556} & 0.3655 \footnotesize{$\pm$0.0225} \\
    \bottomrule
  \end{tabular}
\end{table}

\begin{table}[h]
  \caption{Comparison results on dataset group \textit{HISTORY-SZ32-STP16-NDRP-STD0.5}}
  \centering
  \begin{tabular}{lccccc}
    \toprule
    Model & Hits@1 ($\%$) & Hits@5 ($\%$) & Hits@10 ($\%$) & MR & MRR \\
    \midrule
    RW-Stationary & 11.72 \footnotesize{$\pm$2.02} & 41.07 \footnotesize{$\pm$3.66} & 54.65 \footnotesize{$\pm$4.07} & 34.876 \footnotesize{$\pm$1.670} & 0.2547 \footnotesize{$\pm$0.0231} \\
    RW-Dynamic & 12.25 \footnotesize{$\pm$1.39} & 46.04 \footnotesize{$\pm$3.81} & 64.03 \footnotesize{$\pm$3.78} & 20.600 \footnotesize{$\pm$2.735} & 0.2820 \footnotesize{$\pm$0.0191} \\
    \midrule
    FullGN & 13.79 \footnotesize{$\pm$1.65} & 51.04 \footnotesize{$\pm$3.10} & \textbf{69.30} \footnotesize{$\pm$2.87} & 31.874 \footnotesize{$\pm$4.526} & 0.3051 \footnotesize{$\pm$0.0207} \\
    FullGN-NoAct & 13.90 \footnotesize{$\pm$1.29} & 49.87 \footnotesize{$\pm$2.94} & 67.39 \footnotesize{$\pm$2.87} & \underline{\textbf{15.716}} \footnotesize{$\pm$1.772} & 0.3059 \footnotesize{$\pm$0.0161} \\
    FullGN-Mul & 12.93 \footnotesize{$\pm$0.90} & 45.60 \footnotesize{$\pm$2.75} & 61.88 \footnotesize{$\pm$2.12} & 21.881 \footnotesize{$\pm$2.319} & 0.2807 \footnotesize{$\pm$0.0120} \\
    FullGN-MulMlp & \underline{\textbf{14.89}} \footnotesize{$\pm$0.96} & \textbf{51.20} \footnotesize{$\pm$2.36} & 68.48 \footnotesize{$\pm$2.88} & 17.917 \footnotesize{$\pm$2.680} & \textbf{0.3145} \footnotesize{$\pm$0.0103} \\
    \midrule
    GGNN & \textbf{14.83} \footnotesize{$\pm$1.27} & \underline{\textbf{53.28}} \footnotesize{$\pm$3.22} & \underline{\textbf{70.76}} \footnotesize{$\pm$2.63} & 33.821 \footnotesize{$\pm$7.460} & \underline{\textbf{0.3217}} \footnotesize{$\pm$0.0153} \\
    GGNN-NoAct & 13.84 \footnotesize{$\pm$1.21} & 48.51 \footnotesize{$\pm$2.39} & 65.86 \footnotesize{$\pm$2.34} & \textbf{20.799} \footnotesize{$\pm$1.546} & 0.2957 \footnotesize{$\pm$0.0126} \\
    GGNN-Mul & 14.16 \footnotesize{$\pm$1.06} & 47.27 \footnotesize{$\pm$3.53} & 63.62 \footnotesize{$\pm$3.58} & 23.457 \footnotesize{$\pm$2.224} & 0.2971 \footnotesize{$\pm$0.0168} \\
    GGNN-MulMlp & 14.80 \footnotesize{$\pm$1.26} & 49.47 \footnotesize{$\pm$2.74} & 66.35 \footnotesize{$\pm$3.00} & 22.230 \footnotesize{$\pm$2.410} & 0.3053 \footnotesize{$\pm$0.0142} \\
    \midrule
    GAT & 10.80 \footnotesize{$\pm$1.14} & 43.44 \footnotesize{$\pm$2.81} & 60.12 \footnotesize{$\pm$3.54} & 73.118 \footnotesize{$\pm$7.201} & 0.2538 \footnotesize{$\pm$0.0105} \\
    GAT-NoAct & 12.60 \footnotesize{$\pm$1.48} & 45.84 \footnotesize{$\pm$3.56} & 62.84 \footnotesize{$\pm$2.95} & \textbf{21.903} \footnotesize{$\pm$1.581} & 0.2829 \footnotesize{$\pm$0.0179} \\
    GAT-Mul & 13.49 \footnotesize{$\pm$1.28} & 46.23 \footnotesize{$\pm$3.15} & 62.68 \footnotesize{$\pm$2.85} & 23.887 \footnotesize{$\pm$2.012} & 0.2885 \footnotesize{$\pm$0.0149} \\
    GAT-MulMlp & \textbf{13.75} \footnotesize{$\pm$1.19} & \textbf{47.93} \footnotesize{$\pm$2.72} & \textbf{64.23} \footnotesize{$\pm$2.23} & 22.535 \footnotesize{$\pm$1.538} & \textbf{0.2933} \footnotesize{$\pm$0.0132} \\
    \bottomrule
  \end{tabular}
\end{table}

\begin{table}[h]
  \caption{Comparison results on dataset groups \textit{\{LINE,SINE,LOCATION,HISTORY\}-SZ32-STP16-EDRP-STD0.2}}
  \centering
  \begin{tabular}{llccccc}
    \toprule
    & Model & Hits@1 ($\%$) & Hits@5 ($\%$) & Hits@10 ($\%$) & MR & MRR \\
    \midrule
    \multirow{4}{*}{\footnotesize{LINE}} & \footnotesize{FullGN} & 17.59 \footnotesize{$\pm$1.17} & 65.27 \footnotesize{$\pm$2.71} & \textbf{85.47} \footnotesize{$\pm$1.96} & 8.239 \footnotesize{$\pm$1.769} & 0.3780 \footnotesize{$\pm$0.0141} \\
    & \footnotesize{-NoAct} & 19.39 \footnotesize{$\pm$1.14} & 64.42 \footnotesize{$\pm$2.06} & 83.84 \footnotesize{$\pm$1.67} & 7.189 \footnotesize{$\pm$0.458} & 0.3887 \footnotesize{$\pm$0.0132} \\
    & \footnotesize{-Mul} & 20.20 \footnotesize{$\pm$1.28} & 65.51 \footnotesize{$\pm$2.77} & 84.56 \footnotesize{$\pm$2.30} & 6.260 \footnotesize{$\pm$0.499} & 0.3985 \footnotesize{$\pm$0.0145} \\
    & \footnotesize{-MulMlp} & \textbf{20.26} \footnotesize{$\pm$1.26} & \textbf{66.35} \footnotesize{$\pm$2.87} & 84.88 \footnotesize{$\pm$2.40} & \textbf{6.185} \footnotesize{$\pm$0.501} & \textbf{0.3999} \footnotesize{$\pm$0.0161} \\
    \midrule
    \multirow{4}{*}{\footnotesize{SINE}} & \footnotesize{FullGN} & \textbf{63.08} \footnotesize{$\pm$2.03} & \textbf{90.32} \footnotesize{$\pm$2.38} & \textbf{93.16} \footnotesize{$\pm$2.21} & 18.052 \footnotesize{$\pm$5.502} & \textbf{0.7464} \footnotesize{$\pm$0.0187} \\
    & \footnotesize{-NoAct} & 52.87 \footnotesize{$\pm$2.48} & 72.77 \footnotesize{$\pm$4.75} & 81.82 \footnotesize{$\pm$3.51} & 10.496 \footnotesize{$\pm$2.357} & 0.6193 \footnotesize{$\pm$0.0247} \\
    & \footnotesize{-Mul} & 58.74 \footnotesize{$\pm$2.14} & 75.56 \footnotesize{$\pm$4.72} & 83.33 \footnotesize{$\pm$3.38} & \textbf{7.518} \footnotesize{$\pm$1.472} & 0.6681 \footnotesize{$\pm$0.0281} \\
    & \footnotesize{-MulMlp} & 56.94 \footnotesize{$\pm$1.79} & 76.97 \footnotesize{$\pm$2.38} & 83.37 \footnotesize{$\pm$2.73} & 7.710 \footnotesize{$\pm$1.647} & 0.6606 \footnotesize{$\pm$0.0162} \\
    \midrule
    \multirow{4}{*}{\footnotesize{LOCA}} & \footnotesize{FullGN} & 22.74 \footnotesize{$\pm$5.22} & 68.86 \footnotesize{$\pm$7.39} & 81.88 \footnotesize{$\pm$4.49} & 20.447 \footnotesize{$\pm$6.473} & 0.4213 \footnotesize{$\pm$0.0554} \\
    & \footnotesize{-NoAct} & 41.18 \footnotesize{$\pm$2.98} & 82.45 \footnotesize{$\pm$1.74} & \textbf{91.31} \footnotesize{$\pm$1.61} & \textbf{5.390} \footnotesize{$\pm$0.849} & 0.5881 \footnotesize{$\pm$0.0249} \\
    & \footnotesize{-Mul} & 37.69 \footnotesize{$\pm$4.07} & 77.94 \footnotesize{$\pm$4.28} & 87.52 \footnotesize{$\pm$2.72} & 8.093 \footnotesize{$\pm$1.841} & 0.5501 \footnotesize{$\pm$0.0401} \\
    & \footnotesize{-MulMlp} & \textbf{43.39} \footnotesize{$\pm$3.26} & \textbf{84.32} \footnotesize{$\pm$3.50} & 90.11 \footnotesize{$\pm$3.15} & 6.249 \footnotesize{$\pm$1.747} & \textbf{0.6081} \footnotesize{$\pm$0.0297} \\
    \midrule
    \multirow{4}{*}{\footnotesize{HIST}} & \footnotesize{FullGN} & 18.62 \footnotesize{$\pm$2.96} & 61.12 \footnotesize{$\pm$3.74} & 74.94 \footnotesize{$\pm$3.95} & 32.250 \footnotesize{$\pm$12.497} & 0.3648 \footnotesize{$\pm$0.0264} \\
    & \footnotesize{-NoAct} & 27.96 \footnotesize{$\pm$2.04} & \textbf{65.73} \footnotesize{$\pm$3.95} & \textbf{76.76} \footnotesize{$\pm$3.44} & \textbf{13.564} \footnotesize{$\pm$3.033} & \textbf{0.4494} \footnotesize{$\pm$0.0200} \\
    & \footnotesize{-Mul} & 28.40 \footnotesize{$\pm$3.76} & 61.72 \footnotesize{$\pm$5.35} & 72.94 \footnotesize{$\pm$3.07} & 19.180 \footnotesize{$\pm$3.192} & 0.4332 \footnotesize{$\pm$0.0338} \\
    & \footnotesize{-MulMlp} & \textbf{29.13} \footnotesize{$\pm$2.91} & 62.26 \footnotesize{$\pm$3.71} & 72.31 \footnotesize{$\pm$3.65} & 18.469 \footnotesize{$\pm$2.971} & 0.4416 \footnotesize{$\pm$0.0297} \\
    \bottomrule
  \end{tabular}
\end{table}

\begin{table}[h]
  \caption{Comparison results on dataset groups \textit{\{LINE,SINE,LOCATION,HISTORY\}-SZ32-STP16-EDRP-STD0.5}}
  \centering
  \begin{tabular}{llccccc}
    \toprule
    & Model & Hits@1 ($\%$) & Hits@5 ($\%$) & Hits@10 ($\%$) & MR & MRR \\
    \midrule
    \multirow{4}{*}{\footnotesize{LINE}} & \footnotesize{FullGN} & 16.87 \footnotesize{$\pm$1.85} & \textbf{63.11} \footnotesize{$\pm$2.29} & \textbf{82.59} \footnotesize{$\pm$1.53} & 9.296 \footnotesize{$\pm$1.374} & 0.3669 \footnotesize{$\pm$0.0177} \\
    & \footnotesize{-NoAct} & 18.75 \footnotesize{$\pm$1.67} & 62.33 \footnotesize{$\pm$2.36} & 81.39 \footnotesize{$\pm$2.00} & 8.564 \footnotesize{$\pm$0.822} & 0.3764 \footnotesize{$\pm$0.0189} \\
    & \footnotesize{-Mul} & 19.42 \footnotesize{$\pm$1.79} & 63.04 \footnotesize{$\pm$2.25} & 81.63 \footnotesize{$\pm$1.75} & \textbf{7.811} \footnotesize{$\pm$0.684} & 0.3857 \footnotesize{$\pm$0.0193} \\
    & \footnotesize{-MulMlp} & \textbf{19.63} \footnotesize{$\pm$1.85} & 63.03 \footnotesize{$\pm$2.29} & 81.56 \footnotesize{$\pm$1.59} & 8.029 \footnotesize{$\pm$0.526} & \textbf{0.3867} \footnotesize{$\pm$0.0198} \\
    \midrule
    \multirow{4}{*}{\footnotesize{SINE}} & \footnotesize{FullGN} & 19.08 \footnotesize{$\pm$1.11} & \textbf{62.90} \footnotesize{$\pm$1.61} & \textbf{80.16} \footnotesize{$\pm$1.61} & 18.303 \footnotesize{$\pm$9.662} & 0.3759 \footnotesize{$\pm$0.0117} \\
    & \footnotesize{-NoAct} & 20.41 \footnotesize{$\pm$1.61} & 56.18 \footnotesize{$\pm$2.94} & 73.19 \footnotesize{$\pm$3.01} & 11.914 \footnotesize{$\pm$1.874} & 0.3656 \footnotesize{$\pm$0.0160} \\
    & \footnotesize{-Mul} & 21.53 \footnotesize{$\pm$1.42} & 59.52 \footnotesize{$\pm$2.30} & 75.77 \footnotesize{$\pm$1.73} & \textbf{11.893} \footnotesize{$\pm$1.124} & 0.3829 \footnotesize{$\pm$0.0132} \\
    & \footnotesize{-MulMlp} & \textbf{22.12} \footnotesize{$\pm$1.21} & 60.09 \footnotesize{$\pm$2.66} & 75.74 \footnotesize{$\pm$1.96} & 12.262 \footnotesize{$\pm$1.532} & \textbf{0.3884} \footnotesize{$\pm$0.0134} \\
    \midrule
    \multirow{4}{*}{\footnotesize{LOCA}} & \footnotesize{FullGN} & 15.95 \footnotesize{$\pm$1.08} & 56.61 \footnotesize{$\pm$2.39} & 76.61 \footnotesize{$\pm$1.77} & 15.732 \footnotesize{$\pm$4.803} & 0.3392 \footnotesize{$\pm$0.0100} \\
    & \footnotesize{-NoAct} & 18.55 \footnotesize{$\pm$1.54} & 59.81 \footnotesize{$\pm$2.67} & 78.41 \footnotesize{$\pm$2.44} & 9.751 \footnotesize{$\pm$1.260} & 0.3703 \footnotesize{$\pm$0.0183} \\
    & \footnotesize{-Mul} & 19.32 \footnotesize{$\pm$1.14} & 60.34 \footnotesize{$\pm$1.82} & 79.69 \footnotesize{$\pm$1.92} & 10.382 \footnotesize{$\pm$1.586} & 0.3759 \footnotesize{$\pm$0.0136} \\
    & \footnotesize{-MulMlp} & \textbf{19.36} \footnotesize{$\pm$1.21} & \textbf{60.92} \footnotesize{$\pm$2.14} & \textbf{80.31} \footnotesize{$\pm$1.84} & \textbf{9.352} \footnotesize{$\pm$1.572} & \textbf{0.3783} \footnotesize{$\pm$0.0132} \\
    \midrule
    \multirow{4}{*}{\footnotesize{HIST}} & \footnotesize{FullGN} & 14.00 \footnotesize{$\pm$1.52} & 51.60 \footnotesize{$\pm$2.51} & 70.76 \footnotesize{$\pm$3.17} & 25.511 \footnotesize{$\pm$6.818} & 0.3091 \footnotesize{$\pm$0.0145} \\
    & \footnotesize{-NoAct} & 16.61 \footnotesize{$\pm$1.49} & \textbf{54.18} \footnotesize{$\pm$2.26} & \textbf{70.87} \footnotesize{$\pm$2.95} & \textbf{13.968} \footnotesize{$\pm$1.772} & \textbf{0.3343} \footnotesize{$\pm$0.0149} \\
    & \footnotesize{-Mul} & \textbf{16.97} \footnotesize{$\pm$0.94} & 50.46 \footnotesize{$\pm$2.38} & 67.69 \footnotesize{$\pm$1.77} & 16.933 \footnotesize{$\pm$2.255} & 0.3252 \footnotesize{$\pm$0.0123} \\
    & \footnotesize{-MulMlp} & 16.27 \footnotesize{$\pm$1.31} & 51.63 \footnotesize{$\pm$3.22} & 68.57 \footnotesize{$\pm$3.77} & 16.647 \footnotesize{$\pm$2.403} & 0.3258 \footnotesize{$\pm$0.0158} \\
    \bottomrule
  \end{tabular}
\end{table}

\begin{table}[h]
  \caption{Comparison results on dataset groups \textit{\{LINE,SINE,LOCATION,HISTORY\}-SZ64-STP32-NDRP-STD0.2}}
  \centering
  \begin{tabular}{llccccc}
    \toprule
    & Model & Hits@1 \scriptsize{($\%$)} & Hits@5 \scriptsize{($\%$)} & Hits@10 \scriptsize{($\%$)} & MR & MRR \\
    \midrule
    \multirow{2}{*}{\footnotesize{LINE}} & \footnotesize{GGNN} & 12.12 \footnotesize{$\pm$0.72} & 48.33 \footnotesize{$\pm$1.56} & 72.83 \footnotesize{$\pm$1.45} & 22.322 \footnotesize{$\pm$13.982} & 0.2887 \footnotesize{$\pm$0.0082} \\
    & \footnotesize{-MulMlp} & \textbf{15.54} \footnotesize{$\pm$0.24} & \textbf{54.61} \footnotesize{$\pm$0.89} & \textbf{75.02} \footnotesize{$\pm$0.89} & \textbf{8.863} \footnotesize{$\pm$0.337} & \textbf{0.3324} \footnotesize{$\pm$0.0040} \\
    \midrule
    \multirow{2}{*}{\footnotesize{SINE}} & \footnotesize{GGNN} & 32.11 \footnotesize{$\pm$3.73} & 65.53 \footnotesize{$\pm$9.09} & \textbf{81.73} \footnotesize{$\pm$5.49} & 52.180 \footnotesize{$\pm$23.395} & 0.4681 \footnotesize{$\pm$0.0494} \\
    & \footnotesize{-MulMlp} & \textbf{38.21} \footnotesize{$\pm$6.94} & \textbf{66.82} \footnotesize{$\pm$4.65} & 76.14 \footnotesize{$\pm$3.60} & \textbf{14.249} \footnotesize{$\pm$2.623} & \textbf{0.5177} \footnotesize{$\pm$0.0591} \\
    \midrule
    \multirow{2}{*}{\footnotesize{LOCA}} & \footnotesize{GGNN} & 22.59 \footnotesize{$\pm$4.28} & 72.54 \footnotesize{$\pm$8.18} & 89.27 \footnotesize{$\pm$4.67} & 14.619 \footnotesize{$\pm$5.860} & 0.4335 \footnotesize{$\pm$0.0515} \\
    & \footnotesize{-MulMlp} & \textbf{44.01} \footnotesize{$\pm$2.60} & \textbf{88.96} \footnotesize{$\pm$2.30} & \textbf{95.94} \footnotesize{$\pm$1.27} & \textbf{4.162} \footnotesize{$\pm$0.978} & \textbf{0.6286} \footnotesize{$\pm$0.0226} \\
    \midrule
    \multirow{2}{*}{\footnotesize{HIST}} & \footnotesize{GGNN} & 9.99 \footnotesize{$\pm$5.08} & 38.16 \footnotesize{$\pm$18.57} & 56.51 \footnotesize{$\pm$26.89} & 354.254 \footnotesize{$\pm$620.045} & 0.2299 \footnotesize{$\pm$0.1091} \\
    & \footnotesize{-MulMlp} & \textbf{24.54} \footnotesize{$\pm$1.77} & \textbf{57.98} \footnotesize{$\pm$2.52} & \textbf{68.24} \footnotesize{$\pm$2.39} & \textbf{49.187} \footnotesize{$\pm$10.726} & \textbf{0.3931} \footnotesize{$\pm$0.0205} \\
    \bottomrule
  \end{tabular}
\end{table}

\begin{table}[h]
  \caption{Comparison results on dataset groups \textit{\{LINE,SINE,LOCATION,HISTORY\}-SZ64-STP32-NDRP-STD0.5}}
  \centering
  \begin{tabular}{llccccc}
    \toprule
    & Model & Hits@1 \scriptsize{($\%$)} & Hits@5 \scriptsize{($\%$)} & Hits@10 \scriptsize{($\%$)} & MR & MRR \\
    \midrule
    \multirow{2}{*}{\footnotesize{LINE}} & \footnotesize{GGNN} & 11.66 \footnotesize{$\pm$0.70} & 47.09 \footnotesize{$\pm$1.62} & 71.45 \footnotesize{$\pm$1.10} & 22.898 \footnotesize{$\pm$10.416} & 0.2814 \footnotesize{$\pm$0.0082} \\
    & \footnotesize{-MulMlp} & \textbf{14.95} \footnotesize{$\pm$0.50} & \textbf{52.94} \footnotesize{$\pm$1.55} & \textbf{73.01} \footnotesize{$\pm$1.19} & \textbf{10.949} \footnotesize{$\pm$0.683} & \textbf{0.3231} \footnotesize{$\pm$0.0070} \\
    \midrule
    \multirow{2}{*}{\footnotesize{SINE}} & \footnotesize{GGNN} & 10.47 \footnotesize{$\pm$0.99} & 39.78 \footnotesize{$\pm$4.60} & 61.11 \footnotesize{$\pm$5.50} & 29.445 \footnotesize{$\pm$12.722} & 0.2484 \footnotesize{$\pm$0.0209} \\
    & \footnotesize{-MulMlp} & \textbf{16.68} \footnotesize{$\pm$0.59} & \textbf{50.91} \footnotesize{$\pm$1.88} & \textbf{67.42} \footnotesize{$\pm$1.59} & \textbf{17.379} \footnotesize{$\pm$1.724} & \textbf{0.3238} \footnotesize{$\pm$0.0099} \\
    \midrule
    \multirow{2}{*}{\footnotesize{LOCA}} & \footnotesize{GGNN} & 11.80 \footnotesize{$\pm$1.37} & 45.93 \footnotesize{$\pm$4.26} & 69.08 \footnotesize{$\pm$5.12} & 26.319 \footnotesize{$\pm$8.972} & 0.2784 \footnotesize{$\pm$0.0224} \\
    & \footnotesize{-MulMlp} & \textbf{18.51} \footnotesize{$\pm$0.51} & \textbf{62.85} \footnotesize{$\pm$1.29} & \textbf{82.41} \footnotesize{$\pm$1.07} & \textbf{8.423} \footnotesize{$\pm$0.738} & \textbf{0.3778} \footnotesize{$\pm$0.0059} \\
    \midrule
    \multirow{2}{*}{\footnotesize{HIST}} & \footnotesize{GGNN} & 7.65 \footnotesize{$\pm$0.98} & 32.67 \footnotesize{$\pm$3.84} & 53.00 \footnotesize{$\pm$5.57} & 92.365 \footnotesize{$\pm$19.973} & 0.2064 \footnotesize{$\pm$0.0192} \\
    & \footnotesize{-MulMlp} & \textbf{13.47} \footnotesize{$\pm$0.74} & \textbf{45.79} \footnotesize{$\pm$2.11} & \textbf{62.85} \footnotesize{$\pm$2.35} & \textbf{39.793} \footnotesize{$\pm$5.507} & \textbf{0.2839} \footnotesize{$\pm$0.0125} \\
    \bottomrule
  \end{tabular}
\end{table}

\subsection{More Visualization Results}

\begin{figure}[h]
\centering
\includegraphics[width=1.1\textwidth]{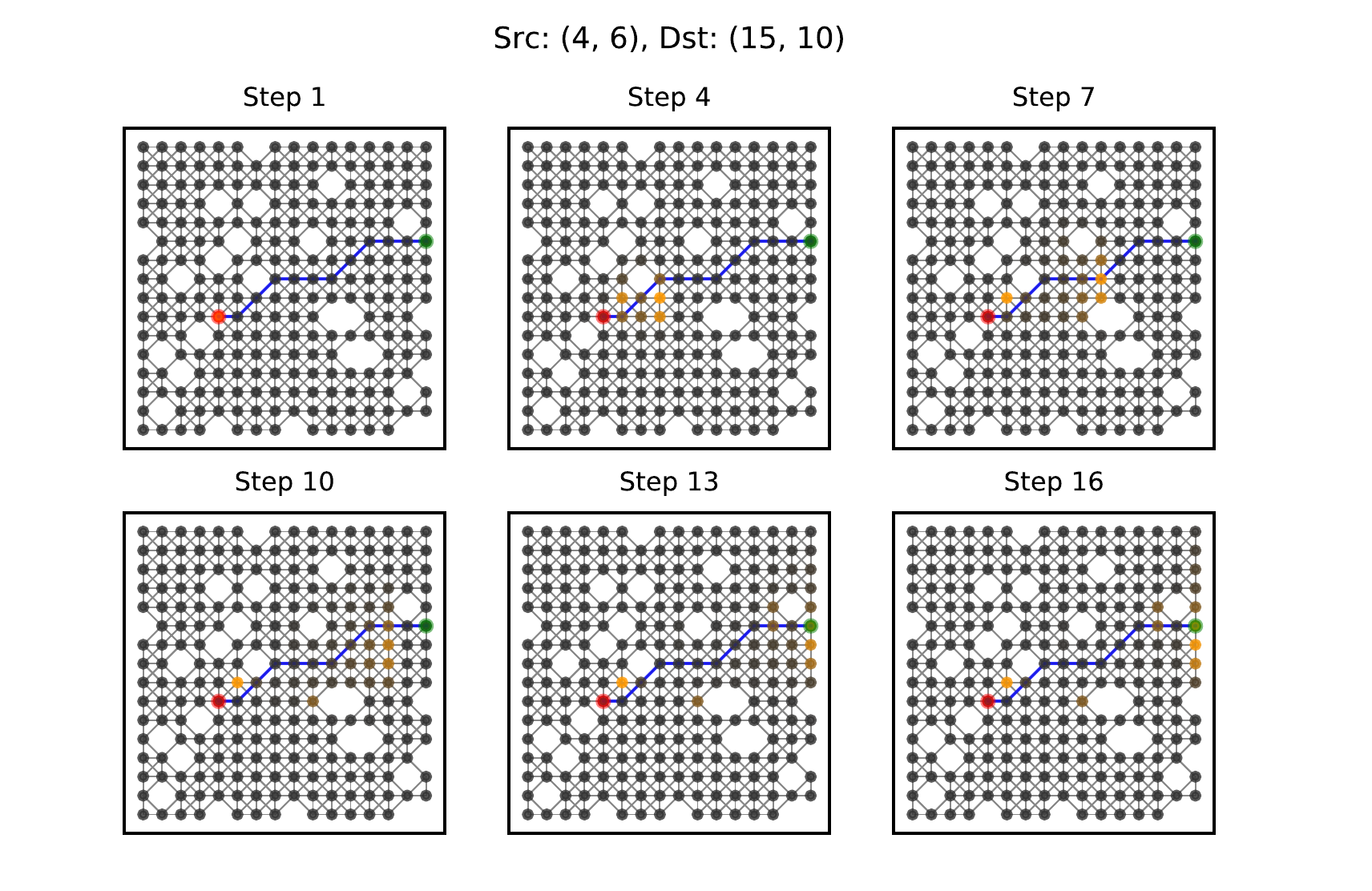}
\caption{Attention distributions at different steps. The latent directions follow a straight line.}
\end{figure}

\begin{figure}[h]
\centering
\includegraphics[width=1.1\textwidth]{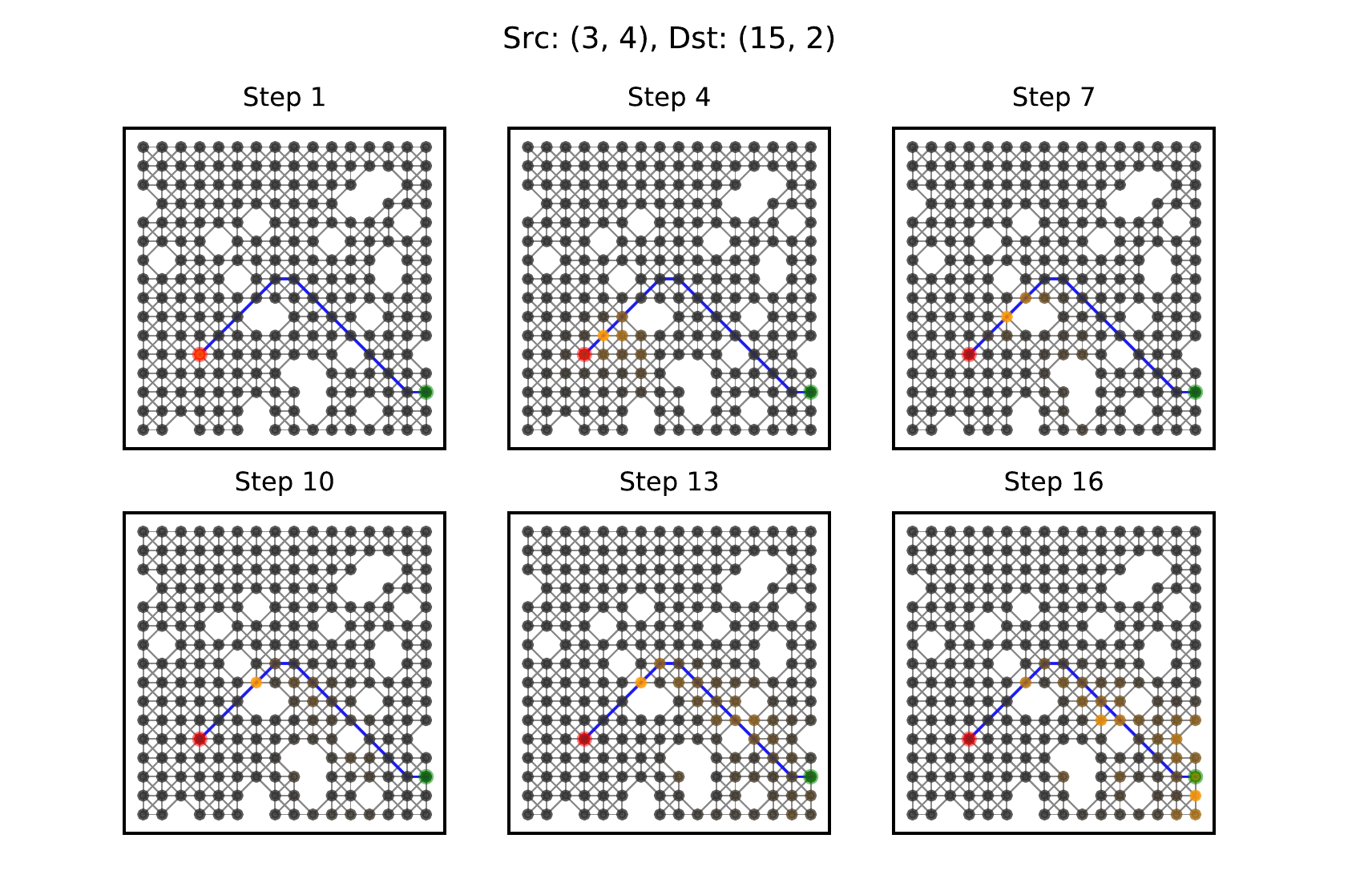}
\caption{Attention distributions at different steps. The latent directions follow a sine curve, depending on time.}
\end{figure}

\begin{figure}[h]
\centering
\includegraphics[width=1.1\textwidth]{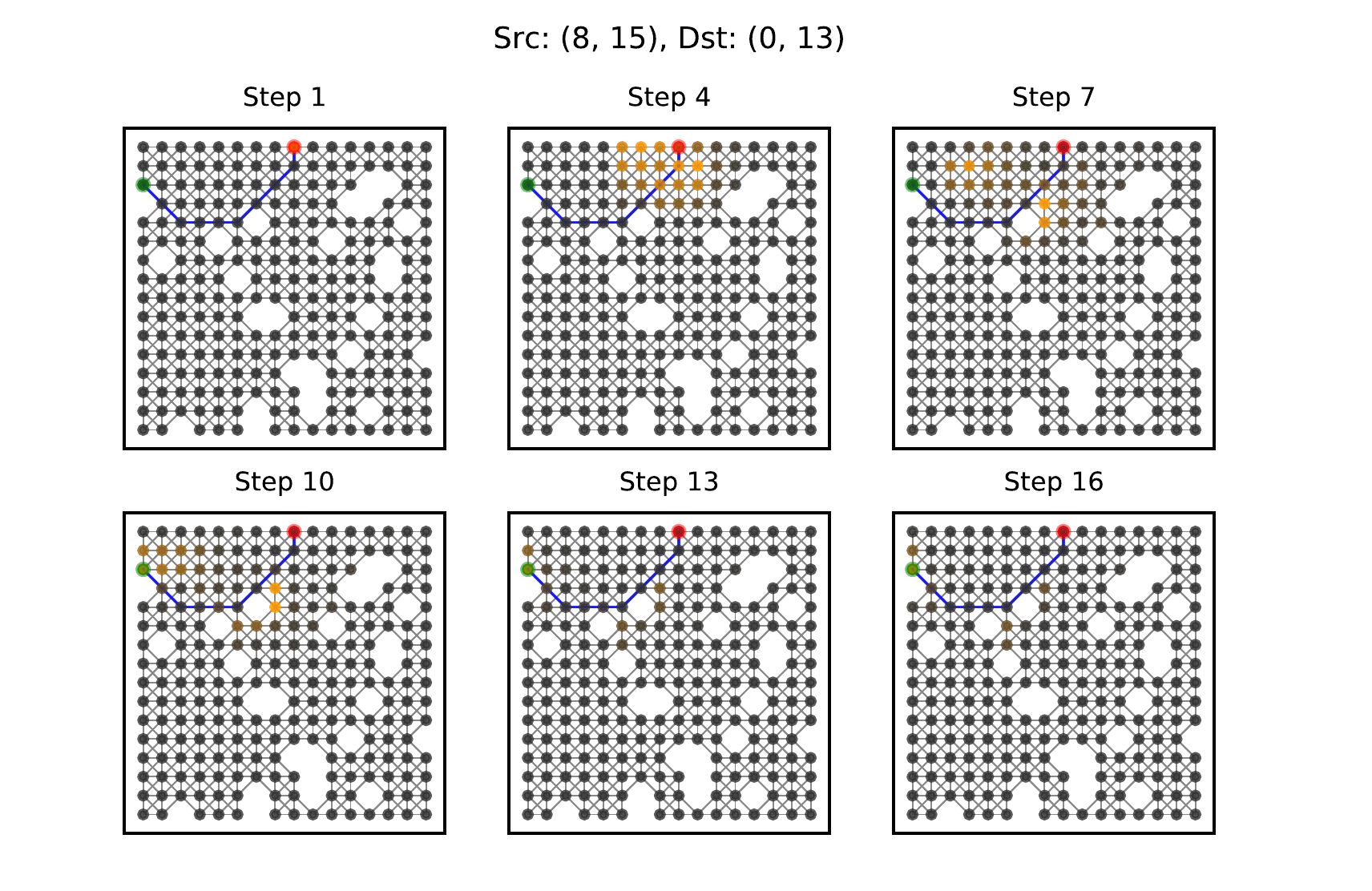}
\caption{Attention distributions at different steps. The latent directions depend on current locations.}
\end{figure}

\begin{figure}[h]
\centering
\includegraphics[width=1.1\textwidth]{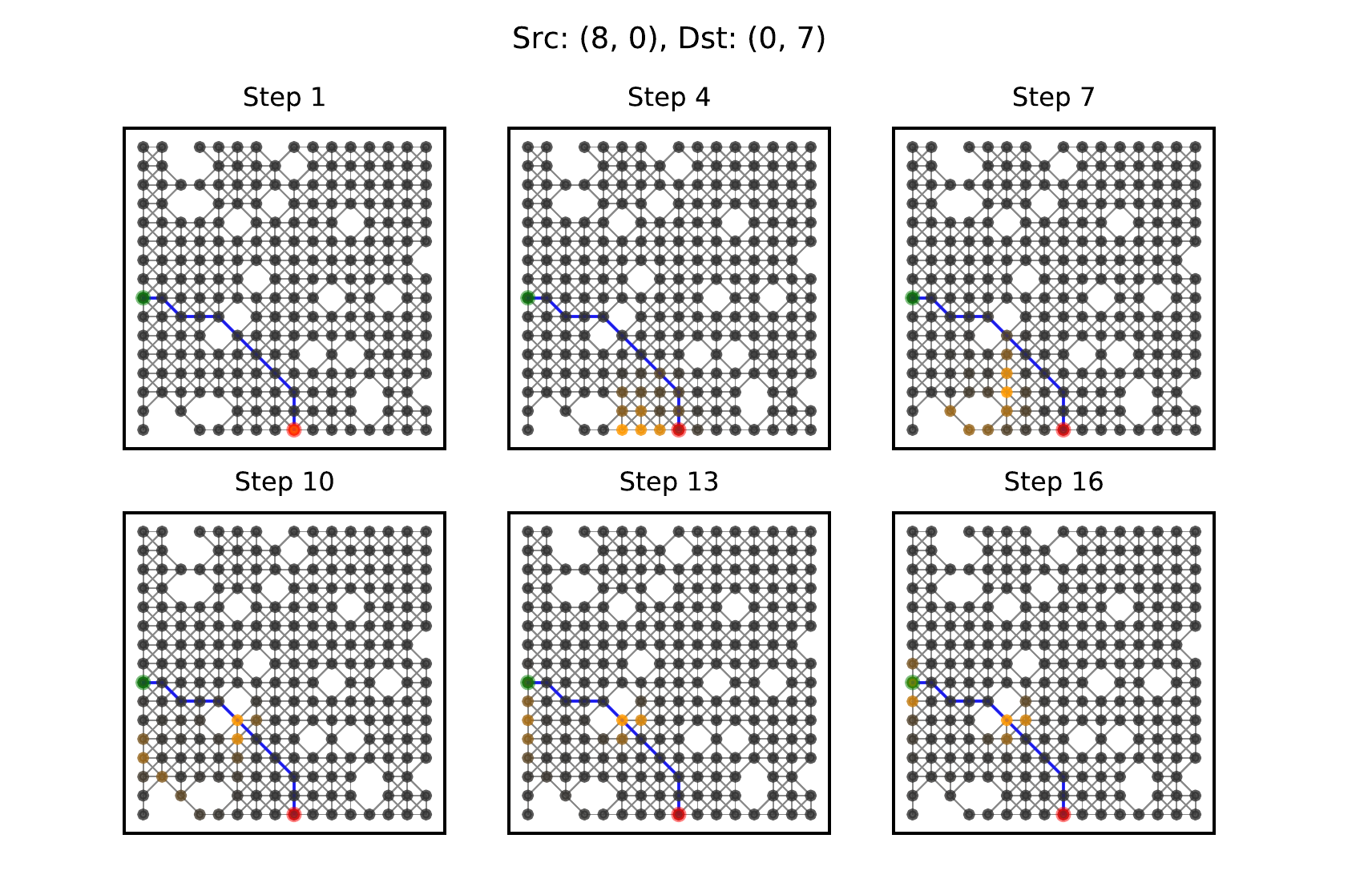}
\caption{Attention distributions at different steps. The latent directions depend on location history.}
\end{figure}

\end{document}